\documentclass[lettersize,journal]{IEEEtran}
\usepackage{amsmath,amsfonts}
\usepackage{algorithmic}
\usepackage{algorithm}
\usepackage{array}
\usepackage{textcomp}
\usepackage{stfloats}
\usepackage{url}
\usepackage{verbatim}
\usepackage{graphicx}
\usepackage{cite}
\usepackage{tabularx}
\usepackage{url}
\usepackage{float}
\usepackage{subcaption}
\usepackage{multirow}
\usepackage{booktabs}
\usepackage{siunitx}
\usepackage[inline]{enumitem}
\usepackage{tabularx}
\usepackage{hyperref}
\hypersetup{colorlinks,allcolors=black}

\newcommand{\mat}[1]{\boldsymbol{#1}}
\renewcommand{\vec}[1]{\boldsymbol{#1}}

\usepackage{fancyhdr}
\fancypagestyle{mahmood}{
  \fancyhf{} % clear all fields
  
  \fancyhead[C]{\footnotesize \textcopyright This work has been submitted to the IEEE for
possible publication. Copyright may be transferred without notice, after
which this version may no longer be accessible.}
}
\makeatletter
\let\ps@IEEEtitlepagestyle\ps@mahmood
\makeatother

\begin{document}

\title{Ultra-Lightweight Collaborative Mapping for Robot Swarms}

\author{Vlad~Niculescu,
        Tommaso~Polonelli,~\IEEEmembership{Member,~IEEE,} Michele~Magno,~\IEEEmembership{Senior Member,~IEEE,}
        and~Luca~Benini,~\IEEEmembership{Fellow,~IEEE}%
        % <-this % stops a space
\thanks{This work is supported in part by BRAINSEE project (8003528831) funded by Armasuisse Science and Technology of the Swiss Confederation. Moreover, it is also partially supported by the ESA – Open Space Innovation Platform (OSIP) under ID: I-2023-03429.}}

% \markboth{Journal of \LaTeX\ Class Files,~Vol.~14, No.~8, August~2024}%


\maketitle

\begin{abstract}
A key requirement in robotics is the ability to simultaneously self-localize and map a previously unknown environment, relying primarily on onboard sensing and computation. Achieving fully onboard accurate simultaneous localization and mapping (SLAM) is feasible for high-end robotic platforms, whereas small and inexpensive robots face challenges due to constrained hardware, therefore frequently resorting to external infrastructure for sensing and computation. 
The challenge is further exacerbated in swarms of robots, where coordination, scalability, and latency are crucial concerns.  
This work introduces a decentralized and lightweight collaborative SLAM approach that enables mapping on virtually any robot, even those equipped with low-cost hardware and only \qty[detect-weight=true, detect-family=true, mode=text]{1.5}{\mega\byte} of memory, including miniaturized insect-size devices. 
Moreover, the proposed solution supports large swarm formations with the capability to coordinate hundreds of agents. 
To substantiate our claims, we have successfully implemented collaborative SLAM on centimeter-size drones weighing \qty[detect-weight=true, detect-family=true, mode=text]{46}{\gram}. 
Remarkably, we achieve a mapping accuracy below \qty[detect-weight=true, detect-family=true, mode=text]{30}{\centi\meter}, a result comparable to high-end state-of-the-art solutions while reducing the cost, memory, and computation requirements by two orders of magnitude. 
Our approach is innovative in three main aspects. 
First, it enables onboard infrastructure-less collaborative mapping with a lightweight and cost-effective (\$20) solution in terms of sensing and computation. 
Second, we optimize the data traffic within the swarm to support hundreds of cooperative agents using standard wireless protocols such as ultra-wideband (UWB), Bluetooth, or WiFi. 
Last, we implement a distributed swarm coordination policy to decrease mapping latency and enhance accuracy. 
\end{abstract}


\section*{Supplementary Material}
Supplementary video at \url{https://youtu.be/uh-Iys90agU} \\ Project's code at \url{https://github.com/ETH-PBL/Nano-C-SLAM}

\section{Introduction} \label{sec:introduction}

Nowadays, swarms of autonomous robots find applications in many sectors, from industry to civil markets, including biomedical and healthcare~\cite{yang2021survey}. 
Key tasks such as perception or mapping can be carried out more effectively and at lower latency by a swarm than by a single agent~\cite{zhou2023racer}. 
However, the design of a collaboration scheme between the agents of a swarm is still an unsolved challenge in many robotics applications~\cite{xu2024d,dorigo2020reflections}. 
The core principle of swarm coordination relies on a set of shared rules based on local-global sensory inputs and communication with neighboring agents. 
In addition, not relying on centralized processing is crucial for increasing robustness, ensuring that the failure of a single robot does not compromise the entire mission execution~\cite{dorigo2020reflections, placed2023survey, lajoie2023swarm}. 
Moreover, achieving autonomous and coordinated robot navigation in real-world environments poses significant scalability challenges in designing a shared coordination scheme that is feasible for hundreds of cooperative agents~\cite{mcguire2019minimal}.

\begin{figure} [t]
\centering
%
%\begin{subfigure}{\columnwidth}
\includegraphics[width=\columnwidth]{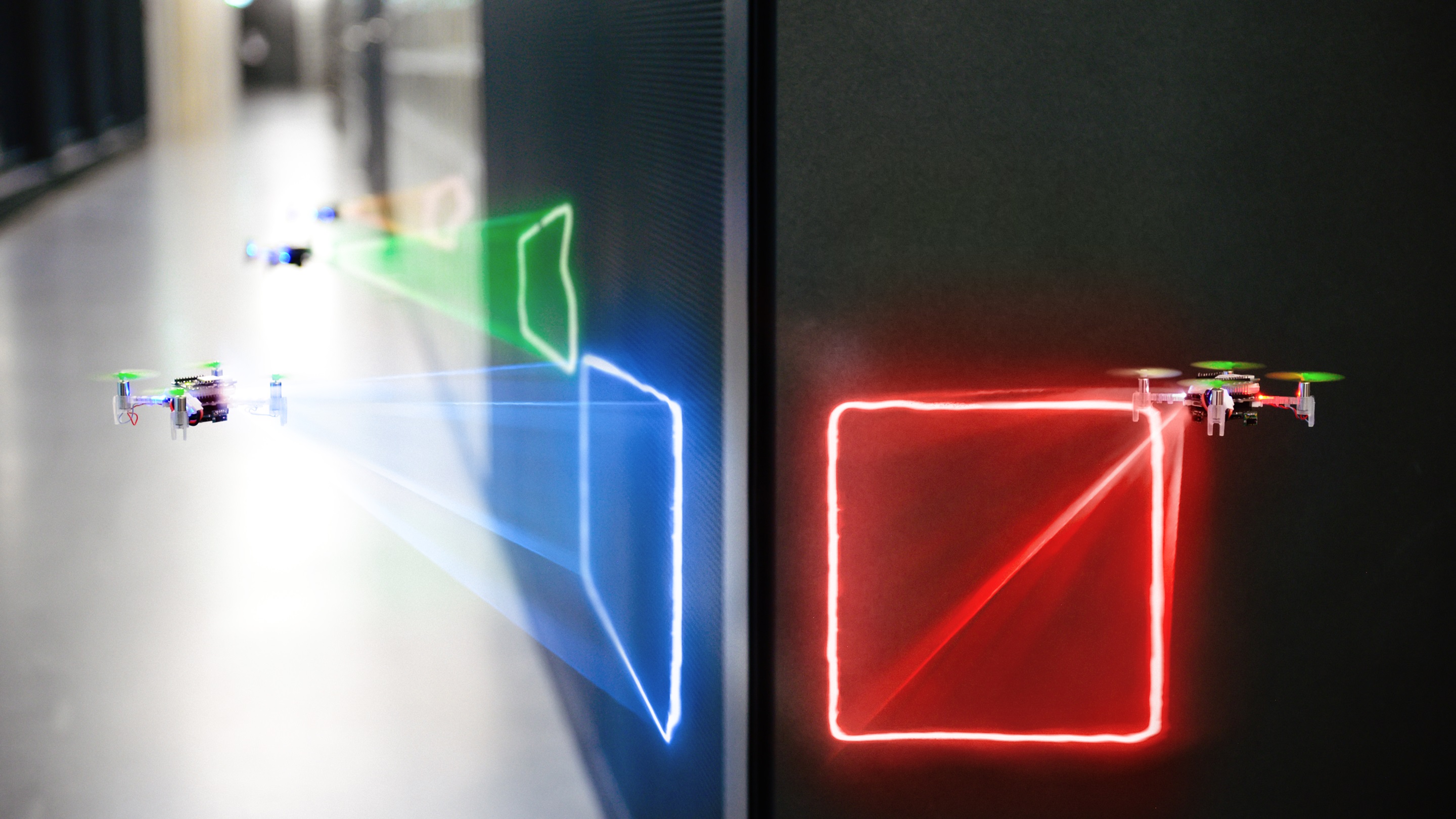}
% \caption{Top view of the four decks.
% \label{fig:laser}}
% \end{subfigure}
% %
% \begin{subfigure}{0.49\columnwidth}
% \centering
% \includegraphics[width=1\linewidth]{figures/single.jpg}
% \caption{Top view of the four decks.
% \label{fig:single}}
% \end{subfigure}
% %
% \begin{subfigure}{0.49\columnwidth}
% \centering
% \includegraphics[width=1\linewidth]{figures/swarm.png}
% \caption{Top view of the four decks.
% \label{fig:swarm}}
% \end{subfigure}
% %
\caption{Illustration of a swarm of miniaturized UAVs using our collaborative SLAM system to map a real environment.}
\label{fig:cover}
\end{figure}

\subsection{Technical Challenges for Infrastructureless Robot Swarms }\label{sec:intro:challenges}
Aerial light shows featuring hundreds of quadcopters and the concept of highly automated warehouses predominantly relying on robotic sorters underscore the current technological landscape~\cite{soria2021predictive, varadharajan2020swarm, amala2023drone}. 
However, those systems mostly follow preprogrammed trajectories calculated on a base station~\cite{kratky2021autonomous}, often relying on external infrastructure for computation, localization, and communication~\cite{mcguire2019minimal, tian2023resilient}. 
Navigation and cooperation of a swarm of robots in unexplored and GNSS-denied environments, without the support of any external framework, remains an open research challenge~\cite{dorigo2020reflections, de2022insect}. 
In particular, the design of an efficient and robust infrastructure-free and decentralized solution for agile collaborative robots, supporting navigation, distributed sensing, and mapping is a challenging research goal in the robotics field~\cite{zhou2022swarm, zhou2021ego, tian2022kimera, lajoie2020door}. 
The ongoing industrial and social revolution is driving the deployment of robots in everyday life, beyond heavily controlled environments, such as automated production lines~\cite{dupont2021decade}. 
Therefore, heterogeneous and cost-effective robotic platforms will soon need to operate in a variety of environments, often uncontrolled, in proximity to humans, animals, and within residential spaces~\cite{arm2023scientific, dorigo2020reflections}. 
In this context, enabling the use of robots across various applications, reducing the hardware costs associated with sensing and computation, is crucial~\cite{iyer2020wireless}. 
Undoubtedly, the current societal viewpoint of autonomous and intelligent machines as exclusive and costly needs to shift towards seeing them as accessible, plug-and-play, and cost-effective tools~\cite{grau2020robots}. 
Moreover, in the context of specific application scenarios, such as space exploration and first-aid, hundreds of droids will be deployed in unexplored and infrastructure-less scenarios without the possibility of human intervention~\cite{arm2023scientific}, pushing even more the need for lightweight, cost-effective, and distributed solutions for autonomous and heterogeneous swarms of miniaturized robots~\cite{muller2023robust, talamali2021less}.

\subsection{Related Works in Distributed SLAM}\label{sec:intro:realted}
Multi-robot distributed simultaneous localization and mapping (SLAM), in which a group of autonomous agents jointly create and maintain a shared map of the surroundings, is one of the essential elements to enable distributed cooperation and optimal task planning~\cite{lajoie2023swarm, talamali2021less, berlinger2021implicit}. 
Typical SLAM methods produce high-resolution 3D maps~\cite{ebadi2023present}, exploiting vision-based solutions based on stereo cameras or light detection and ranging (LiDAR) depth sensors~\cite{cai2023occupancy}. However, these standard approaches are generally very demanding in terms of computational and memory resources, requiring expensive system-on-module (SoM) platforms integrating CPUs and GPUs with several gigabytes of local memory~\cite{rosinol2020kimera, baumann2024forzaeth}. 
Other limitations include the system cost of the hardware, sensing, and the wireless bandwidth required to exchange real-time information with other agents in the swarm. 
State-of-the-art (SoA) dense vision-based SLAM (vSLAM) approaches require the transfer of up to \qty{3.5}{GB} of data in a centralized scheme, or $\sim$\qty{150}{MB} per robot in a distributed and communication-efficient scheme~\cite{tian2022kimera, huang2022edge}. 
Therefore, an increased number of robots would saturate a WiFi network with only a few agents, in the range between 3 and 10, generating a fundamental scalability bottleneck~\cite{tian2022kimera, huang2022edge}. Overall, there has yet to be a fully distributed and highly scalable SLAM system addressing all the associated challenges of supporting large numbers of collaborative agents operating with limited onboard resources without relying on an external infrastructure for sensing and/or computation. These challenges encompass communication and computational requirements alongside the hardware costs for sensing and data processing~\cite{talamali2021less}.

\subsection{Contribution}\label{sec:intro:contribution}
This work approaches the research challenge of collaborative SLAM (C-SLAM) with a novel approach in every aspect, from sensing to computation, in challenging real-world environments such as indoor areas filled with obstacles. The primary contribution is a fully distributed C-SLAM system for multi-robot dense mapping using sparse sensing. Our system enables a swarm of collaborative robots to estimate a 2D mesh model of the environment in real-time. Each agent runs entirely onboard the SLAM algorithm to process depth-inertial sensor data based on local state estimation and a low-power, cost-effective, and lightweight 64-pixel depth camera. Wireless communication can be supported by any commercial standard, including WiFi and Bluetooth low energy (BLE)~\cite{jamshed2022challenges}, while for the scope of this work we employ ultra-wideband (UWB) to also enable infrastructure-less intra-swarm ranging (i.e., distance estimation). A fully distributed localization and ranging procedure is triggered to perform inter-robot collision avoidance and place recognition. Each swarm agent performs real-time local mesh updates to correct mapping drift through multi-robot loop closure using the iterative closest point (ICP) algorithm~\cite{vizzo2023kiss}, enhancing both local accuracy and global map consistency. Moreover, the proposed implementation is modular, allowing different agents to join/leave the swarm dynamically, supporting heterogeneous robots. 

While LiDARs paired with depth-based SLAM proved to yield high mapping accuracy, they have a high cost of a few hundred dollars. Thus, mounting such a unit onboard every robot in a swarm would dramatically increase the total cost. We demonstrate the possibility of enabling mapping with similar SoA accuracy using \$5 time-of-flight (ToF) $8\times8$ multizone depth sensors and a low-power commercial-off-the-shelf (COTS) microcontroller. So far, SLAM, and especially C-SLAM, has been a prerogative for high-end platforms featuring GPUs coupled with several gigabytes of memory that cost hundreds of dollars. On the other hand, our highly optimized C-SLAM pipeline can run in real-time on COTS ultra-low-power microcontrollers featuring \qty{1.5}{MB} of RAM. With a power budget of $\sim$\qty{100}{\milli\watt} and a cost below \$10, our C-SLAM engine can correct the map in about \qty{250}{\milli\second} after every loop closure. The required power budget, including sensing and computation, is below \qty{1}{\watt}. Therefore, our work provides mapping capabilities to a wide range of robotic platforms, including inexpensive, lightweight, and resource-constrained drones.

In addition to the theoretical formulation and simulation-based studies, our work demonstrates with practical and extensive field experiments how a team of ultra-constrained centimeter-size unmanned aerial vehicles (UAVs) can collaboratively map a generic environment relying only on onboard capabilities for mapping, communication, localization, and sensing. Figure~\ref{fig:cover} illustrates our swarm of nano-UAVs~\cite{muller2023robust} deployed in-field, performing collaborative mapping of an indoor environment. Our contributions can be summarized as follows: \textit{(i)} a distributed lightweight C-SLAM framework that runs onboard and performs depth-based loop closure and distributed trajectory optimization. \textit{(ii)} A communication scheme that enables the robots to coordinate and minimizes the amount of exchanged data. \textit{(iii)} A navigation strategy that attempts to uniformly distribute the swarm in an unexplored environment. \textit{(iV)} The whole project, together with the hardware, the firmware, the navigation policy, and the mathematical formulation, is released open-source.

While our experimental evaluation is performed with UAVs, our C-SLAM system can be deployed on any robot that can accommodate a payload of approximately \qty{10}{\gram}, a power budget of \qty{1}{\watt}, and that is equipped with odometry capabilities. Note that the odometry capabilities can even be achieved using low-cost and low-resolution sensors~\cite{fuller2022gyroscope}, such as the PMW3901 camera that features a resolution of 35 $\times$ 35 pixels. These sensors do not need to provide high accuracy, as the odometry errors are corrected by the C-SLAM algorithm.
\section{Algorithms} \label{sec:algorithms}
In this section, we introduce the ultra-lightweight C-SLAM algorithm and the exploration strategy that the robots use to improve the mapping coverage in unknown environments. Furthermore, we expound upon the C-SLAM scheme and the mechanism used by each swarm agent to rectify its trajectory and align maps to ensure global consistency.
Our solutions can be paired with any robotic platform, as long as it provides depth measurement capabilities enabled by sensors, such as the $8 \times 8$ STMicroelectronics VL53L8CX.

In the following, we assume a system equipped with four VL53L8CX depth sensors, each oriented differently (i.e., front, back, left, right) to provide omnidirectional coverage.
Although the depth ToF sensors output information in matrix format (i.e., depth across vertical and horizontal directions), our system performs 2D mapping and therefore we need to reduce the depth measurements to one plane. For this purpose, we reduce each matrix to a depth row by selecting the median pixel from each column. Given that the sensors employed to demonstrate the effectiveness of our solutions have a resolution of $8\times8$, this translates to 8 pixels after the reduction or a total of 32 pixels for all four sensors.

\subsection{Collaborative SLAM scheme}

Our mapping system employs graph-based SLAM, which represents the robot trajectory as a pose graph. The nodes are poses sampled at discrete times (i.e., $\vec{x}_k$) and the edges are relative measurements between the poses. Each pose $\vec{x}_k=(x_k,y_k,\psi_k)$ is expressed in the world frame and consists of the 2D position and heading, where k represents the timestamp. Furthermore, we note as $\vec{z}\sb{i,j}$ the relative measurement from poses $i$ to $j$. Any two consecutive nodes within the graph of a robot are connected by an edge characterized by the 3-element vector $\vec{z}\sb{i,i+1}$ provided by the drone’s state estimator -- named odometry edge. At every instant $k$, the robot not only stores a new pose but also acquires 32 distance measurements from the four depth sensors, which constitute a depth frame. The poses and their associated depth frames (i.e., sparse measurements) are sufficient to produce the map by projecting the distance measurements in the world frame (dense mapping). 

\begin{figure} [t]
\begin{centering}
\includegraphics[width=\linewidth]{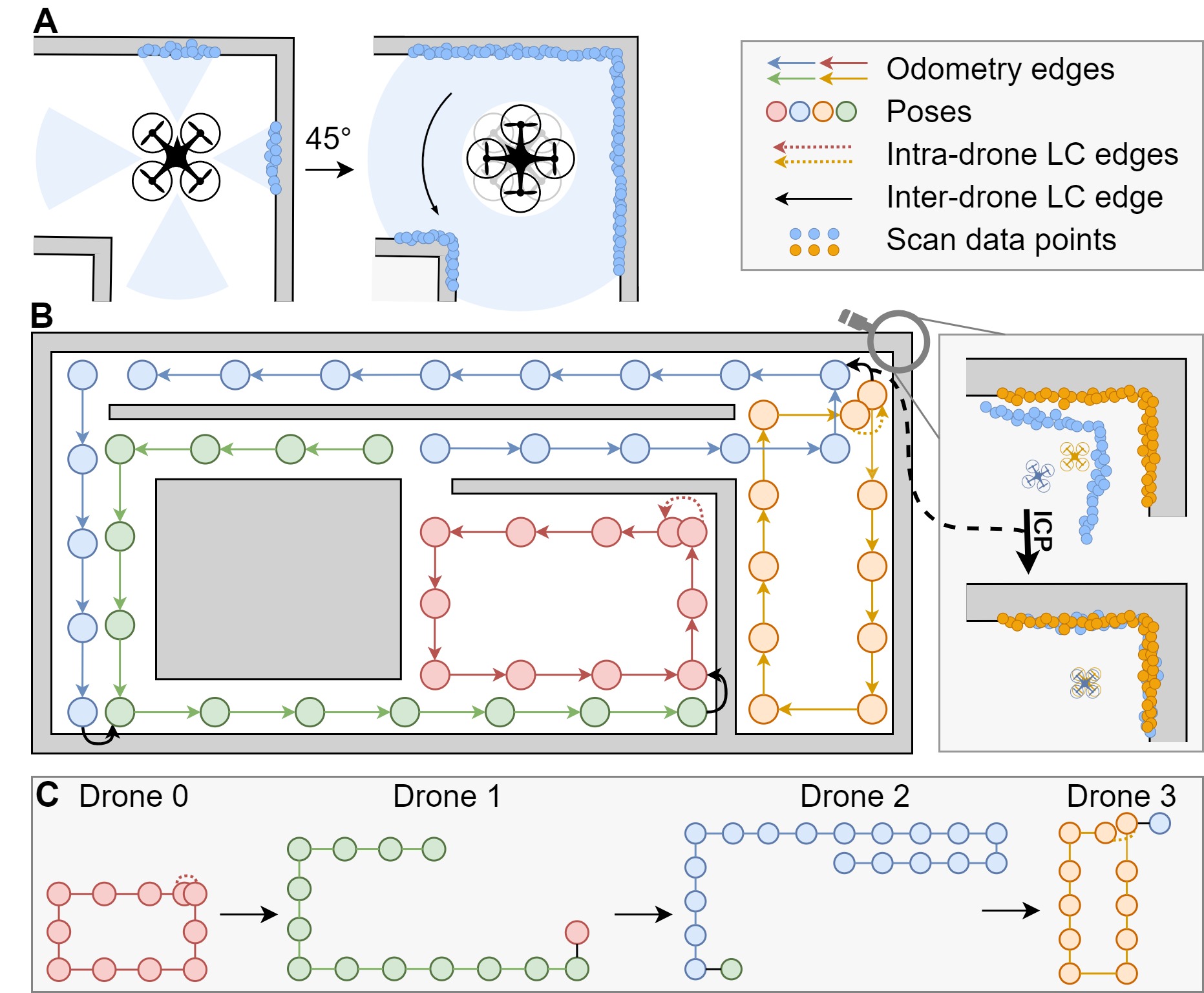}
\par\end{centering}
\centering{}
\caption{The C-SLAM scheme. (A) Illustration of how a drone acquires a scan once reaching a texture-rich location. (B) Composite visualization of individual pose graphs and how they are connected through inter-drone loop closure edges. (C) The cascaded distributed SLAM optimization, illustrating the graph optimized by each drone.}
\label{fig:cslam}
\end{figure}

However, the odometry edges are affected by errors, and computing the pose values using forward integration would cause the trajectory to drift over time. Consequently, our C-SLAM uses a correction mechanism that compares observations of the environment in revisited locations. These observations are tiny map tiles generated out of 20 consecutive depth frames, named \textit{scans}. Figure~\ref{fig:cslam}-A shows how a scan is created while appending projected depth frames. Because the cumulative field of view (FoV) of the four depth sensors is \qty{180}{\degree}, the drone also spins in place by at least \qty{45}{\degree} to achieve full coverage during the scan acquisition. Each scan is matched with a pose, representing the robot position right before the acquisition starts. Using scan-matching, our system determines the optimal rotation and translation that overlaps one scan over the other. Since the poses and their associated scans are acquired at the same time, the transformation resulting from scan-matching also applies to the poses. In this way, when the drone revisits a location, scan-matching determines an accurate rigid body transformation relative to a pose that was previously acquired in that location, to perform loop closure. 

We distinguish two types of loop closures: \textit{(i)} intra-drone loop closures, occurring when the compared scans belong to the same drone, and \textit{(ii)} inter-drone loop closures, occurring when scan-matching ovelaps scans belonging to two different robots. The intra and inter-drone loop closures are essential for ensuring the accuracy of the local-global maps and for maintaining overall consistency on the reference frame.
The information provided by scan-matching is incorporated in the graph as an additional loop closure edge. Then, our system uses pose-graph optimization (PGO)~\cite{niculescu2023nanoslam, fan2023majorization}, propagating the loop closure information through the whole graph to correct all previously acquired poses. In our work, we employ the ICP algorithm to perform scan-matching\cite{vizzo2023kiss}.

Thus, the scan-matching mechanism facilitates not only the correction of a robot's pose in relation to its own previous pose (i.e., intra-drone loop closure) but also in relation to the pose of another drone (i.e., inter-drone loop closure). Figure~\ref{fig:cslam}-B shows an example of the graphs associated with multiple agents, where each agent maps a different environmental area. Nonetheless, an overlapping segment is necessary to establish connections between the drones' respective graphs. 
Within our distributed optimization scheme, each agent exclusively engages in inter-loop closures with drones possessing lower IDs. Furthermore, the graph each drone optimizes consists of its own poses alongside the external poses. To ensure coherence across all graphs, external poses remain unaltered during PGO, serving solely as reference points (i.e., anchor points) for optimizing self poses. In other words, each drone aligns its map relative to the drones possessing lower IDs.

Equation~\ref{eq:cost} shows the optimization problem of the distributed PGO, which is solved onboard by each agent leveraging the approach from~\cite{niculescu2023nanoslam}. 
Let $\mat{X} = \{\vec{x}_0, \vec{x}_1,  \ldots \}$ be the set of all poses in the graph. 
We note as $\hat{\vec{z}}\sb{i,j}$ the prediction of an edge measurement, which is the edge measurement computed out of two poses $\vec{x}_i$ and $\vec{x}_j$. Therefore, solving the problem in Equation~\ref{eq:cost} reduces to determining the pose values that ensure consistency between the edge measurements $\vec{z}\sb{i,j}$ and the predicted measurements $\hat{\vec{z}}\sb{i,j}$. The maximum likelihood solution associated with Equation~\ref{eq:cost} is equivalent to minimizing the sum of squared differences $\vec{e}\sb{i,j} = \vec{z}\sb{i,j} - \hat{\vec{z}}\sb{i,j}$. Each term $\vec{e}\sb{i,j}$ is paired with an edge and weighted by a diagonal information matrix $\Omega$. 
When computing the errors associated with the odometry edges, $j=i+1$ as the odometry edges are always connecting consecutive poses. Moreover, the external poses $\vec{x}_k$ used to compute the inter-drone loop closure edges remain unchanged during the optimization.

\begin{align} \label{eq:cost}
X^* = \arg\min_{x} ( &\sum_{i} \underbrace{e_{i,i+1}^T \Omega_{\text{odom}} e_{i,i+1}}_{\text{odometry edges}} + \sum_{i,j} \underbrace{e_{i,j}^T \Omega_{\text{LC}} e_{i,j}}_{\text{intra LC edges}} + \nonumber\\
+ &\sum_{i,k} \underbrace{e_{i,k}^T \Omega_{\text{LC}} e_{i,k}}_{\text{inter LC edges}} )~.
\end{align}

\subsection{Deriving Inter-Drone Loop Closure Constraints}

\begin{figure} [t]
\begin{centering}
\includegraphics[width=0.7\linewidth]{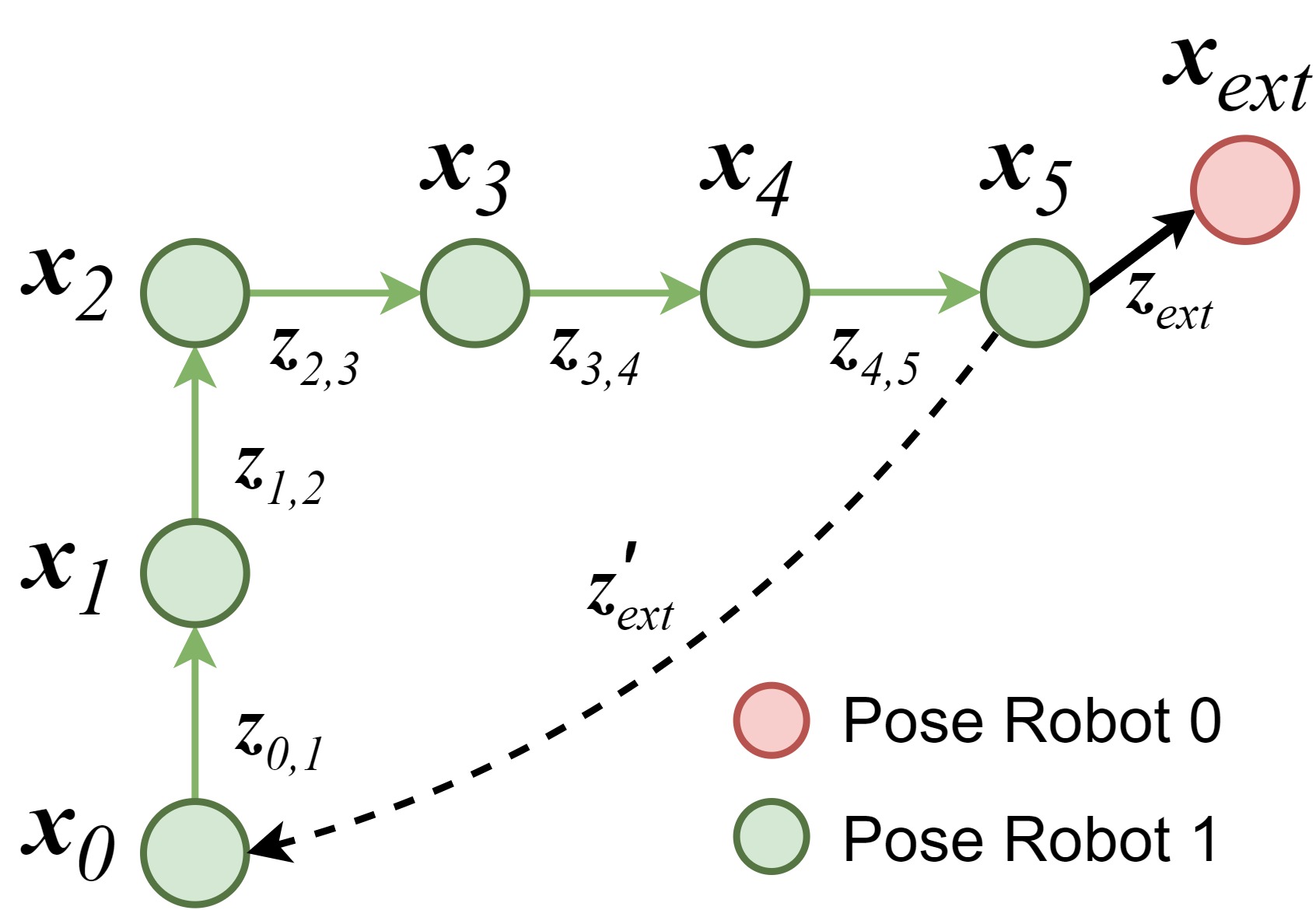}
\par\end{centering}
\centering{}
\caption{An example graph consisted of 6 internal poses and one external pose, showing how an external pose is transformed into an internal constraint.}
\label{fig:small_graph}
\end{figure}

As prerequisites for the optimization, each drone requires the value of the first pose (the take-off position) and the edge values. Moreover, the PGO algorithm never changes the first pose value, which anchors the whole graph. The graph each robot optimizes consists of the poses derived from its own trajectory and the poses from other robots that are involved in loop closure with the respective robot. Figure~\ref{fig:small_graph} shows a simple example of a graph, with 6 poses acquired internally (i.e., directly from the robot) and one external pose (i.e., $\vec{x}\sb{ext}$) involved in loop closure with the pose $\vec{x}_5$. While the odometry edges $\vec{z}\sb{0,1}, \vec{z}_{1,2}, \ldots,  \vec{z}_{4,5}$ are calculated with the aid of the internal state estimator, the inter-drone loop closure edge $\vec{z}\sb{ext}$ is provided by running ICP scan-matching on the scans associated to poses $\vec{x}_5$ and $\vec{x}\sb{ext}$. 

\begin{equation} \label{eq:lc_edge}
    Z_{\text{ext}} = ( \underbrace{Z_{\text{ICP}} X_5}_{X_5^{'}})^{-1} X_{\text{ext}}
\end{equation}

\begin{equation} \label{eq:lc_new_edge}
    Z'_{\text{ext}} = ( \underbrace{Z_{\text{ICP}} X_5}_{X_5^{'}})^{-1} X_0 
\end{equation}

Let $X_i$ be the homogenous coordinates representation of a pose $\vec{x}_i$ and $Z_i$ the homogenous coordinates representation of an edge $\vec{z}_i$. Therefore, if $\vec{z}\sb{ICP}$ is the scan-matching result, then the loop closure edge value is computed as in Equation~\ref{eq:lc_edge}. We recall that the pose $\vec{x}\sb{ext}$ should act as reference for the graph of Robot 1 and correct the value of $\vec{x}\sb{5}$ so that the scans of $\vec{x}\sb{5}$ and $\vec{x}\sb{ext}$ overlap – and therefore also the maps of the two robots. Consequently, $\vec{x}\sb{5}\sp{'}$ (shown in Equation~\ref{eq:lc_edge}) represents the adjusted value of pose $\vec{x}\sb{5}$. However, including the pose $\vec{x}\sb{ext}$ and edge $\vec{z}\sb{ext}$ in the graph would not achieve this result. This is because PGO would change the value of $\vec{x}\sb{ext}$ during optimization and therefore also the value of $\vec{x}\sb{5}$ -- despite maintaining the relative difference between the two poses. To mitigate this issue, we transform the loop closure edge from $\vec{x}\sb{5}$ to $\vec{x}\sb{ext}$ (i.e., $\vec{z}\sb{ext}$) into an edge from $\vec{x}\sb{5}$ to $\vec{x}\sb{0}$ -- which we note as $\vec{z}\sb{ext}\sp{'}$ in Figure~\ref{fig:small_graph}. This exploits the fact that the value of $\vec{x}_0$ never changes during PGO. We therefore provide the relation for computing the new edge $\vec{z}\sb{ext}\sp{'}$ in Equation~\ref{eq:lc_new_edge}.

Note that the structure is very similar to Equation~\ref{eq:lc_edge}, but now the transformation is relative to $\vec{x}_0$ and not to $\vec{x}\sb{ext}$. With this method the external poses exist in the graph only virtually, because any edge from an external pose $\vec{x}\sb{ext}$ to an internal pose $\vec{x}\sb{i}$ is transformed into an edge from $\vec{x}_i$ to $\vec{x}_0$, and $\vec{x}\sb{ext}$ is not included in the graph. 
Therefore, inter-drone loop closure edges do not increase the number of nodes in the graph. This is important because the complexity of optimizing the graph is quadratic in the number of nodes.

\subsection{Update Inter-Constraints}
PGO is a process that typically occurs several times during a mapping mission. This implies that the pose values dynamically change over time. However, since the inter-drone loop closure edges depend on the poses of the drones with lower IDs, it follows that the loop closure edges need to be updated throughout the mission. Considering the graph example in Figure~\ref{fig:small_graph}, imagine a situation where $\vec{z}\sb{ext}\sp{'}$ is computed based on the value of $\vec{x}\sb{ext}$, but then $\vec{x}\sb{ext}$ changes in the next PGO of Robot 0. Thus, before Robot 1 can perform PGO as well, it firstly needs to update $\vec{z}\sb{ext}\sp{'}$. One option would be that Robot 0 sends again the scan to Robot 1, which recomputes $\vec{z}\sb{ICP}$ by running ICP. However, this solution is not practical due to the communication overhead of sending the scan and the computation overhead of re-running ICP. For a large number of robots in the swarm, this would make every optimization extremely slow, posing scalability issues. 
The alternative we propose is to only transmit the updated values of the poses involved in loop closure. In the example presented here, this is equivalent to sending the optimized value of pose $\vec{x}\sb{ext}$ from Robot 0 to Robot 1. We note this value as $\vec{x}\sb{ext\_new}$. The idea is to update the edge value $\vec{z}\sb{ext}\sp{'}$ based on the difference between the old pose value $\vec{x}\sb{ext}$ and the new value $\vec{x}\sb{ext\_new}$ as shown in Equation~\ref{eq:upd_edge}.

\begin{equation} \label{eq:upd_edge}
    Z'_{ext} = \left( X_{ext\_new} X_{ext}^{-1} Z_{ICP} X_5 \right)^{-1} X_0~.
\end{equation}

We highlight that Equation~\ref{eq:upd_edge} is formulated specifically for the example introduced here. However, it generalizes for any inter-drone loop closure edge, computed out of an internal pose $\vec{x}\sb{i}$ and an external pose $\vec{x}\sb{ext}$. Thus, after a robot performs PGO, it also broadcasts the new values of the poses involved in loop closure with other robots. When higher ID robots receive the new pose values, they update the loop closure edges according to Equation~\ref{eq:upd_edge}.

\section{Communication and Ranging Protocol}\label{sec:comm_rang}

\begin{figure} [t]
\begin{centering}
\includegraphics[width=\linewidth]{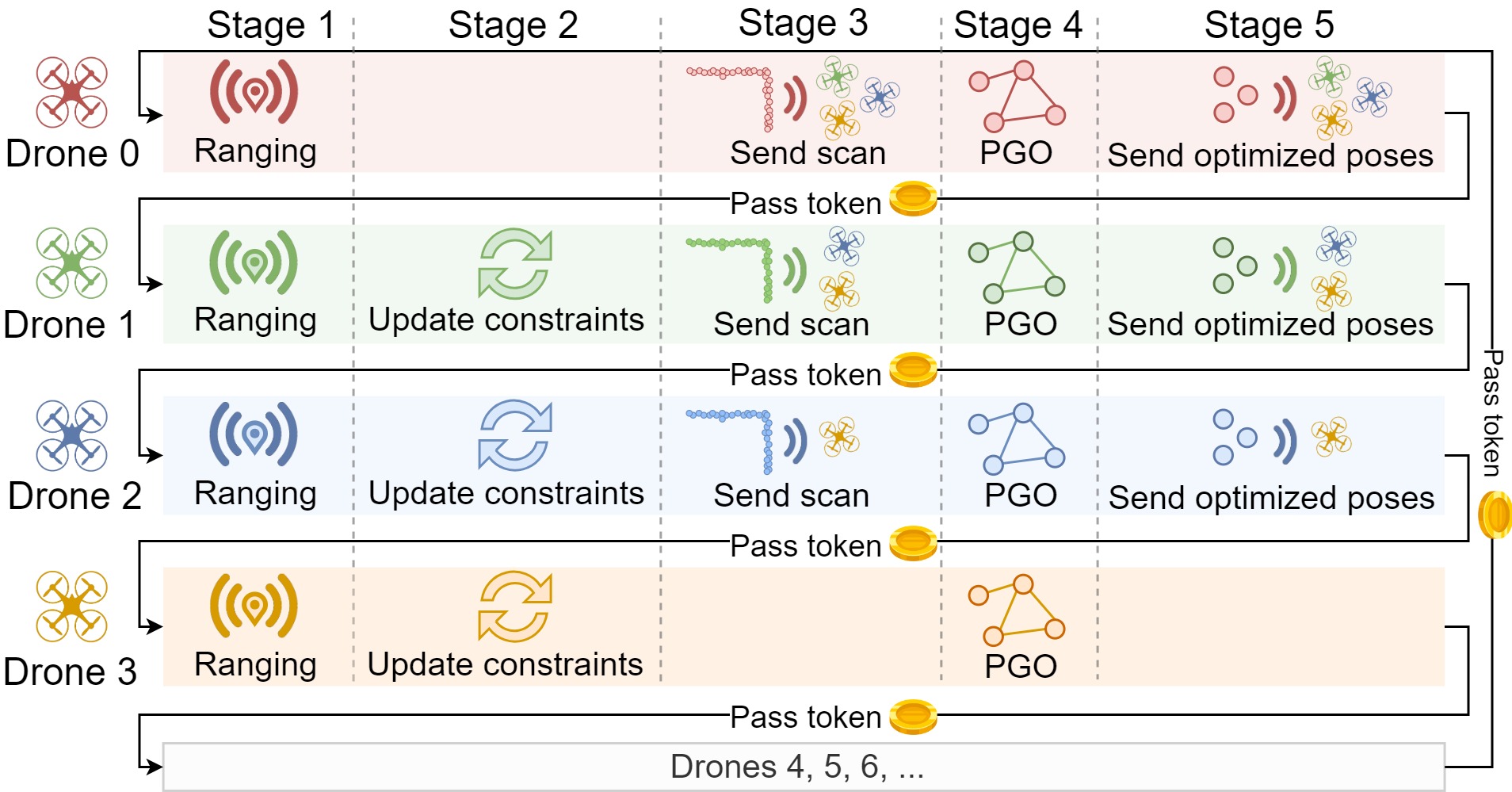}
\par\end{centering}
\centering{}
\caption{The communication protocol, supporting a variable number of swarm agents while minimizing data traffic.}
\label{fig:cslam2}
\end{figure}

The exploration algorithm requires knowledge of the relative positions and distances to all the other swarm agents. Furthermore, the C-SLAM scheme requires scans from the other drones to perform inter-drone loop closures. Figure~\ref{fig:cslam2} shows the multi-stage communication protocol that orchestrates how the drones exchange information. 
The protocol is implemented using UWB for the scope of this work, which is used for both data transmission and ranging. However, it can be implemented leveraging alternative commercial standards, such as WiFi and BLE~\cite{jamshed2022challenges}. Moreover, the protocol is token-based, implying that only one drone can transmit at a time, avoiding packet collision effects and, consequently, decreasing wireless traffic. After a drone receives the token, it first performs ranging with all drones of higher ID. We employ the double-sided two-way-ranging (IEEE 802.15.4a-2007)~\cite{zhou2022swarm, shan2021ultra}, which involves a pairwise point-to-point distance estimation. 
Furthermore, the drones also embed their current positions (from the pose graph) in the ranging messages. After drone $i$ completes Stage 1, it is aware of all distances and positions of the drones with ID $i+1$ or higher. 

In Stage 2, a drone verifies if it has any received external scans in the buffer. 
Then, it pairs each external scan with the internal scan acquired at a distance smaller than a threshold $D\sb{th}=\qty{1}{\meter}$. If multiple internal scans meet this condition for a given external scan, the system chooses the one acquired the earliest. For each pair, it runs ICP and derives a new inter-drone loop closure edge. In Stage 3, the drone checks if a new scan has been acquired since the last time it had the token, and if this is the case, it broadcasts the scan to all drones with higher IDs. The other drones store the received scan in a buffer and process it during Stage 2 the next time they acquire the token. Stage 4 consists of optimizing the pose graph. This is always initialized by drone 0 when it completes a certain number of loop closures or periodically at a defined time interval. To initiate the cascaded PGO, drone 0 sends an extra message to drone 1 along with the token handoff. Subsequently, each drone mirrors this action, relaying the message forward in sync with the token's progression to notify all drones to perform PGO during their next turn. After every PGO, the drone also sends the new values of the scan poses to the other drones with higher IDs – performed within Stage 5. The size of the updated poses (i.e., \qty{12}{\byte} per pose) is negligible compared to the scan size.

We mention that while the communication protocol iterates continuously through the drones in the swarm, it is only Stage 1 which is executed every time the drone acquires the token. An arbitrary drone $i$ requires $N-i-1$ distance measurements in Stage 1 to communicate with all higher ID drones, where N is the total number of drones. This results in a total of at most $N(N-1)/2$ communications within a whole round, where each ranging requires about \qty{4}{\milli\second}. Stage 2 is typically performed several times per minute, depending on how often external scans are received and, therefore, how many texture-rich locations are in the environment. The time spent in this stage depends quasi-linearly on the number of drones and it is dominated by the execution time of ICP (typically below \qty{20}{\milli\second} per scan-matching). Since PGO does not need to happen too frequently, Stages 4 and 5 are typically executed once per minute in a normal indoor scenario.

The designed system exhibits robust fail-safe characteristics, capable of addressing radio disruptions, hardware malfunctions, or temporary absence of swarm agents. All data messages are acknowledged, and in the event of a missing response, retransmission is tried up to three times. In instances where a drone, designated by ID $i$, attempts to pass the token to drone $i+1$ without acknowledgment, there is an automated procedure to redirect the token to drone $i+2$, and so on. Additionally, any drone in the swarm will automatically reclaim the token if more than $2(i+1) s$ have passed since the last time it had the token. This mechanism preserves the functionality of the protocol even when the token-bearing drone falls outside of the communicable range or experiences a failure. 

The representation in Figure~\ref{fig:cslam2} shows only a simplified version of the protocol. However, after each token acquisition, a drone enters a brief listening period, spanning a random interval between 0 -- \qty{20}{\milli\second} to ascertain if another drone transmits during this time (i.e., has the token). Should the transmitting drone have a lower ID, the current drone discards the token. If not, it proceeds with the subsequent stages. This mechanism mitigates the situations where more drones have the token simultaneously. For example, if the drone carrying the token moves out of the swarm's radio range, another drone will claim the token. When the missing drone returns to the swarm, our protocol determines which drone keeps the token.

\begin{figure*} [t]
\begin{centering}
\includegraphics[width=\linewidth]{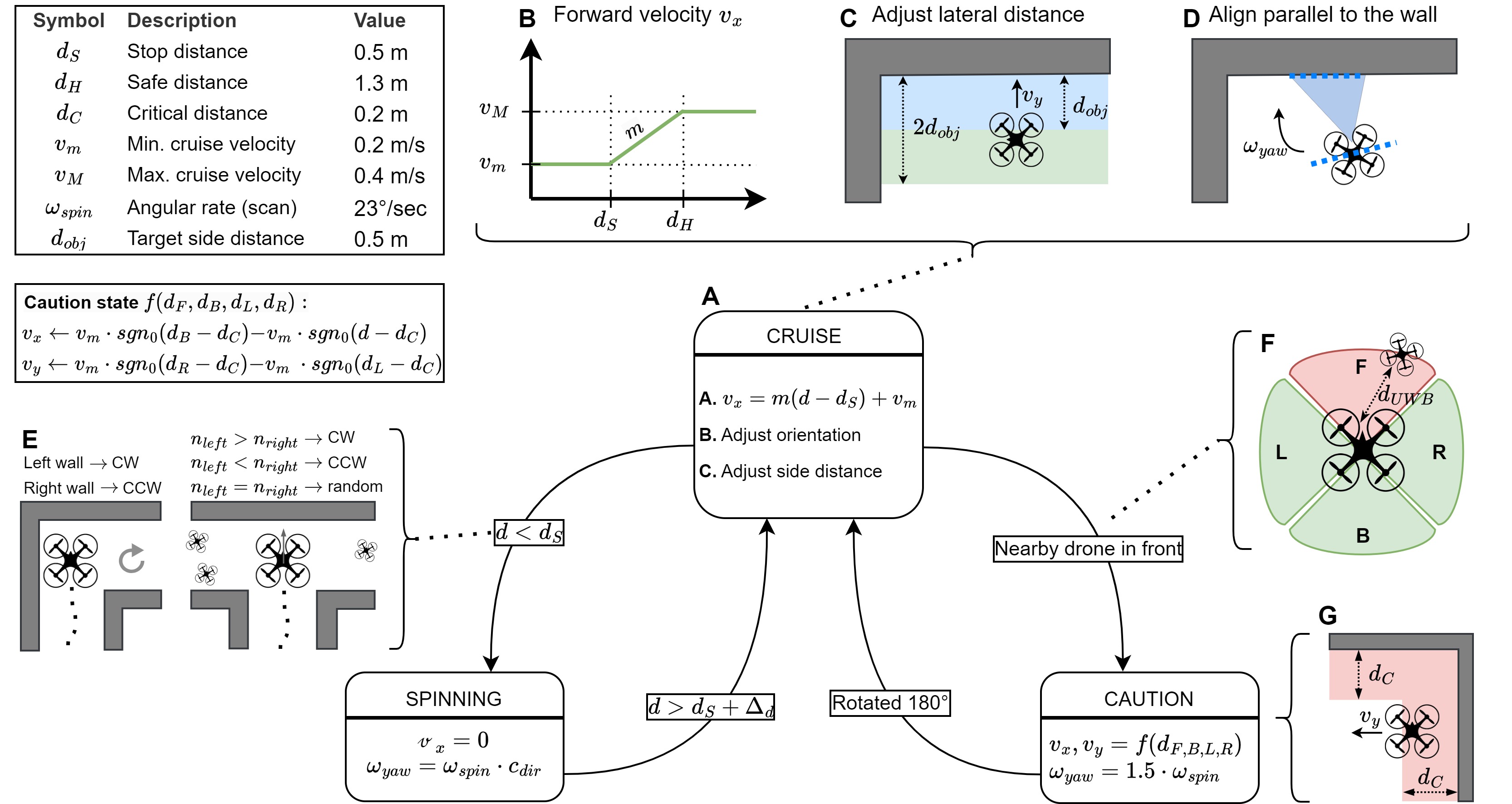}
\par\end{centering}
\centering{}
\caption{The state machine of the exploration algorithm illustrating the behavior in each state and the transition conditions.}
\label{fig:exploration}
\end{figure*}

\section{Exploration Algorithm}\label{sec:exploration}
We introduce our lightweight exploration strategy used for the scope of this paper, which is formulated as a state machine and governs the behavior of individual robots. This strategy performs a pseudo-random exploration of a given environment without requiring a target destination. However, unlike other random exploration algorithms~\cite{mcguire2019minimal}, this strategy prioritizes wall following and corner visiting~\cite{duisterhof2021sniffy}, which encompass texture-rich areas favorable to loop closures. Regardless of the state, the goal of the exploration algorithm is to provide the body velocities (i.e., $v_x$, $v_y$), and the heading angular rate $\omega$. 
Figure~\ref{fig:exploration}-A illustrates the three states of the state machine, where each robot starts in the \textit{Cruise} mode, characterized by a cyclic execution of three steps. In the initial step, the algorithm adjusts the forward velocity $v_x$  linearly, dependent on the frontal distance $d$, clipped with predefined lower and upper bounds $v\sb{min}$ and $v\sb{max}$. Note that distance $d$ is computed as the minimum value provided by the frontal depth sensor. The velocity curve is shown in Figure~\ref{fig:exploration}-B, where m represents the slope of the linear region. 
The second step verifies if there is a side obstacle within a lateral distance of $2d\sb{obj}$. A proportional controller then adjusts $v_y$ to position the drone at $d\sb{obj}$ from the detected side obstacle, as shown in Figure~\ref{fig:exploration}-C. 
Figure~\ref{fig:exploration}-D illustrates the final step of the \textit{Cruise} state, where the robot checks for the presence of an object on either side. A line detector assesses if the depth measurements align with the expected readings from a flat surface. Based on the slope of the line, the system aligns the robot's orientation parallelly. 
If an object is detected on both sides, priority is given to the right. If an object is detected beyond the distance $d\sb{obj}$ or not detected at all, the commanded angular rate $\omega$ is set to 0. The first step of the \textit{Cruise} state ensures smooth deceleration upon detecting a frontal obstacle, while subsequent steps facilitate effective wall following.

When a frontal obstacle is detected, and $d$ becomes smaller than a threshold $d_S$, the system goes into the \textit{Spinning} state. The direction of spinning is determined by evaluating $d_L$ and $d_R$ distances on the left and right, respectively. In the event of an obstacle detected to the left (i.e., $d_L < 2d\sb{obj}$), the robot rotates clockwise (CW) as shown on the left side of Figure~\ref{fig:exploration}-E; conversely, if an obstacle is detected to the right, the robot rotates counter-clockwise (CCW). 
In scenarios where both sides are unobstructed, and the robot is situated at an intersection (i.e., Figure~\ref{fig:exploration}-E – right side), it counts how many other agents are on its left and right relative to its OX-axis (shown in grey) and turns towards the side with fewer robots. This mechanism favors a uniform robot distribution within the environment and increases the area coverage rate. In the event of an equal number of robots on both sides, a random decision is made. When $d$ becomes higher than $d_S + \Delta_d$, with $\Delta_d=\qty{0.3}{\meter}$, a pathway ahead is again clear, and the system makes the transition back to the \textit{Cruise} state. 

If the mission is performed by one drone only, the system would only loop between \textit{Cruise} and \textit{Spinning} states. The transition to the \textit{Caution} state can only be triggered by the interaction with another robot. Figure~\ref{fig:exploration}-F shows how each robot virtually splits the surroundings into four areas. If there is an agent within the frontal (F) area at a distance lower than $2d\sb{obj}$, i.e., $d\sb{UWB} < 2d\sb{obj}$, the system goes into the \textit{Caution} state. In finding the presence of another agent within the frontal zone, the robot utilizes the received position estimates, while for determining proximity within a distance $2d\sb{obj}$, it solely relies on UWB measurements. In the \textit{Caution} state, the robot turns around 180°. During spinning, it also checks if an obstacle (i.e., wall or another agent) is located closer than a critical distance $d_C$, and sets $v_x$ and $v_y$ accordingly to push itself away from the obstacle, as illustrated in Figure~\ref{fig:exploration}-G. Thus, to avoid collisions, when two or more robots encounter one another head-on, they each turn around and continue exploring.
Note that the distances $d_B$,$d_R$, and $d_L$ represent the minimum distances from the back, right, and left areas, respectively. When the robot completes spinning, it returns to the \textit{Cruise} state.

\section{Hardware setup} \label{sec:hardware}

\begin{figure}[t]
    \centering
    \begin{subfigure}{\columnwidth}
    \centering
    \includegraphics[width=1\linewidth]{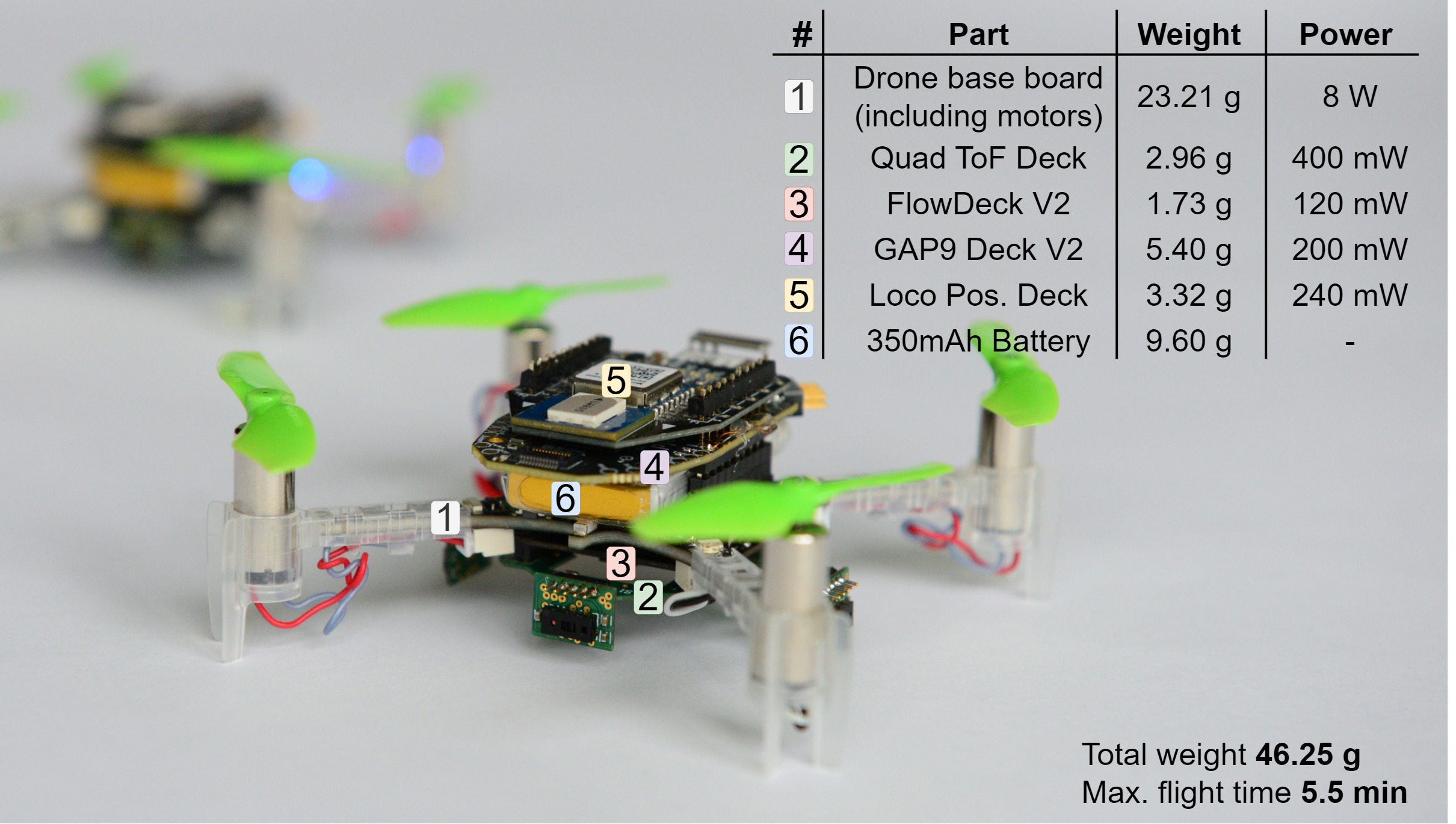}
    \caption{The fully integrated system.
    \label{fig:hardware-a}}
    \end{subfigure}
    \begin{subfigure}{0.49\columnwidth}
    \centering
    \includegraphics[width=1\linewidth]{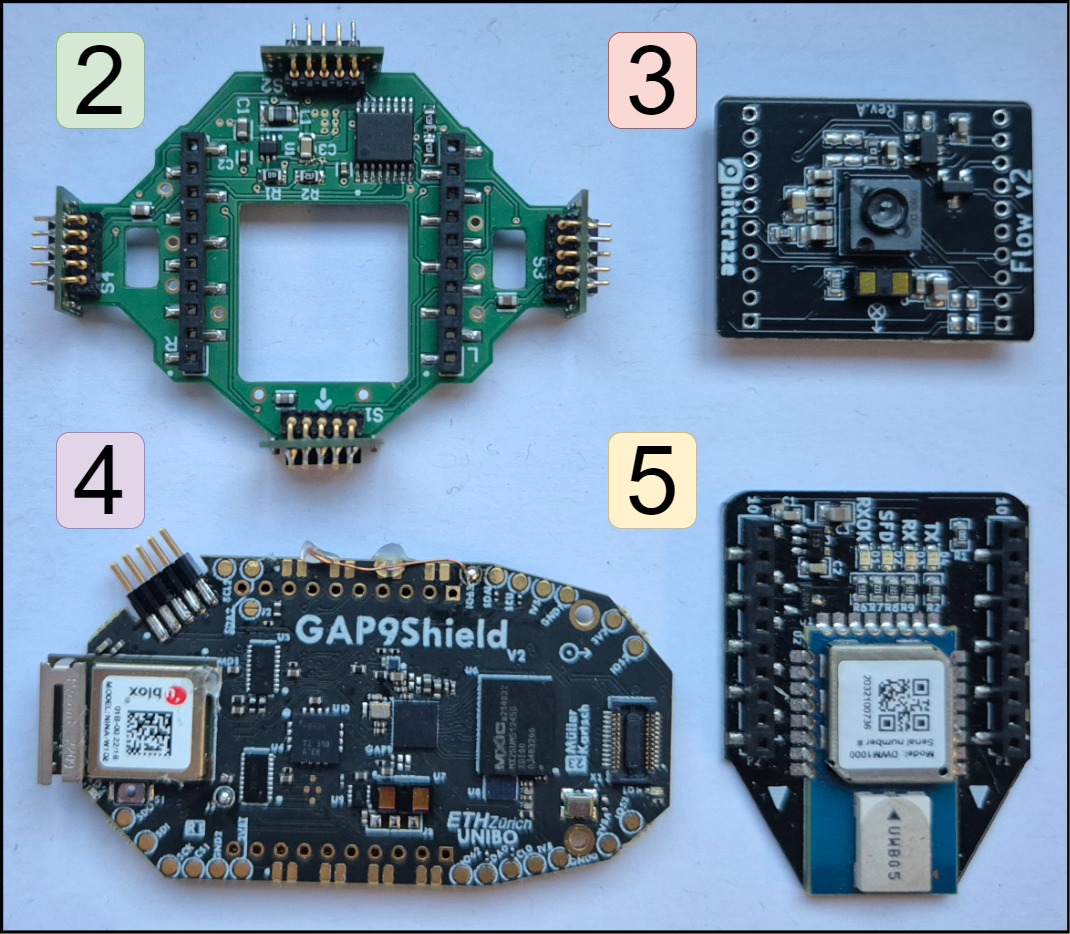}
    \caption{Top view of the four decks.
    \label{fig:hardware-b}}
   \end{subfigure}
   \begin{subfigure}{0.49\columnwidth}
   \centering
   \includegraphics[width=1\linewidth]{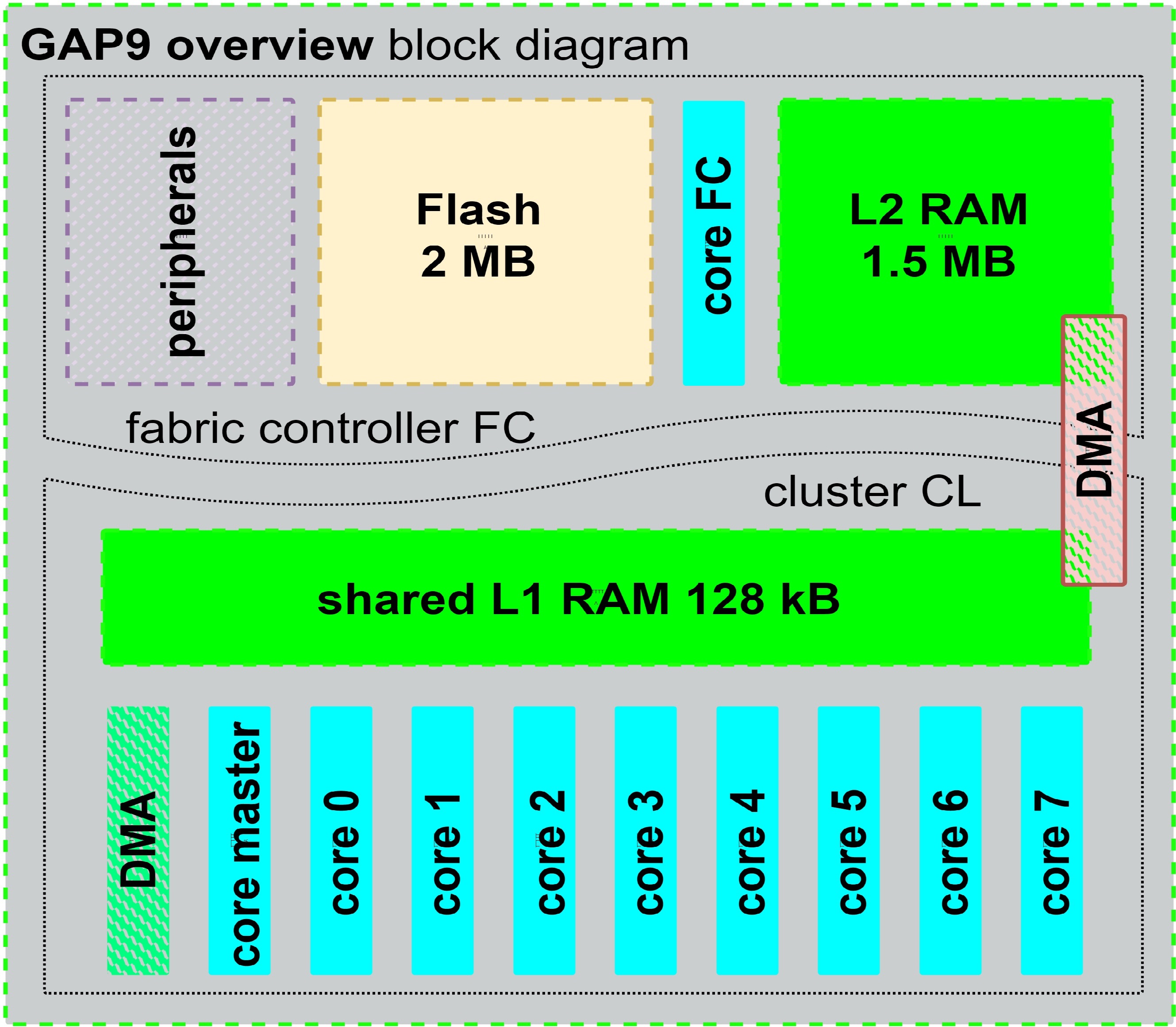}
   \caption{Architecture of the GAP9.
   \label{fig:hardware-c}}
   \end{subfigure}
\caption{(a) The hardware platform used in our experiments, based on the Crazyflie v2 nano-drone. (b) The plug-in decks used by our system, that stack on the base drone PCB. (c) The constitutive blocks of the GAP9 SoC from Greenwaves, showing the 8+1 cluster cores, and the fabric controller. 
\label{fig:hardware}}
\end{figure}

Our mapping system is designed with flexibility in mind, aiming to provide compatibility with a wide range of robotic platforms. The primary requirement is the availability of depth sensors. As a result, the algorithm and implementation can be adjusted to support alternative hardware setups, such as various processors or sensing components. 
For our demonstration, we have chosen the commercial-of-the-shelf (COTS) nano-UAV Crazyflie 2.1 from Bitcraze. This option allows us to demonstrate the effectiveness of our solution even on highly constrained platforms. Consequently, our results can be easily replicated using commercially available hardware.

The open-source firmware of Crazyflie 2.1 offers functionalities for flight control, state estimation, radio communication, and setpoint commander. The drone's primary printed circuit board (PCB) serves also as frame, housing essential electronics such as an inertial measurement unit (IMU), a radio transceiver Nordic nRF51822, and an STM32F405 microcontroller unit (MCU). The latter operates at a maximum clock frequency of \qty{168}{\mega\hertz} and has \qty{192}{kB} of RAM, with over 70\% of these resources already utilized by the firmware for control and estimation tasks. Additionally, the drone is equipped with extension headers for adding extra decks (i.e., plug-in boards). 

We have incorporated the commercial FlowDeck v2, which utilizes a downward-facing optical flow camera and a single-zone ToF ranging sensor. These components enable velocity and height measurements, fused by the onboard extended Kalman filter (EKF) for position and heading estimation. Moreover, we use the Loco Positioning deck featuring IEEE 802.15.4 UWB communication and ranging, via the COTS Decawave DWM1000 module. In addition, we include two custom-designed decks: one housing four depth ToF sensors to enhance the drone's environmental sensing capabilities, and the second deck featuring the GAP9 system-on-chip (SoC). This serves as a co-processor, extending the computational capabilities of Crazyflie 2.1. Therefore, computationally intensive tasks such as running ICP scan-matching, graph-based SLAM, scan computation, or map generation are executed on the GAP9. 
In this configuration, the total weight at take-off is \qty{46.25}{\gram}, which includes all the hardware utilized for the scope of this paper. The fully integrated system, highlighting our custom hardware, is depicted in Figure~\ref{fig:hardware-a}, while Figure~\ref{fig:hardware-b} shows the decks mounted on the drone. The STM32 MCU serves as the manager for all processes running onboard the nano-UAV, including sensor data acquisition. While it does not handle heavy computations itself, it is responsible for delegating these tasks to GAP9 via SPI communication. During the flight, the STM32 retrieves depth data from the four ToF sensors and the current pose from the internal state estimator, transmitting this information to GAP9.

\subsection{The Co-processor Deck Featuring the GAP9}
We use the co-processor deck proposed in~\cite{muller2024gap9shield}, which weighs \qty{5.4}{\gram} and incorporates the GAP9 SoC -- based on the parallel ultra-low-power (PULP) paradigm~\cite{rossi2021vega}, and developed by Greenwaves Technologies. 
Figure~\ref{fig:hardware-c} illustrates the primary components of the GAP9. 
The GAP9 SoC features 10 RISC-V-based cores, organized into two power and frequency domains. 

The first domain is the fabric controller (FC), housing a single core capable of operating at speeds of up to \qty{400}{\mega\hertz}, paired with \qty{1.5}{\mega\byte} of RAM (L2 memory). The FC serves as the main processor of the SoC, managing communication with peripherals and coordinating on-chip memory operations. The second domain is the cluster (CL), comprising nine RISC-V cores capable of operating at speeds of up to \qty{400}{\mega\hertz}. These cores are specifically designed to handle highly parallelizable and computationally intensive workloads. Within the CL, one core acts as a "master core," receiving tasks from the FC and distributing them to the other eight cores in the cluster, which execute the computation. The CL cores share the same instruction cache, enabling efficient execution of identical code on different data. The CL is accompanied by \qty{128}{\kilo\byte} of L1 memory (shared among the CL cores), with transfers between L2 and L1. Moreover, the CL features a neural engine that provides hardware-level acceleration for operations like convolutions, batch normalization, or ReLU activations, supporting the execution of quantized deep learning models. 

The GAP9 interfaces with the STM32 via SPI and handles all the intensive computation required by PGO and scan-matching.

\subsection{Custom Quad ToF Deck}
The VL53L5CX is a lightweight multi-zone 64-pixel ToF sensor, weighing only \qty{42}{\milli\gram}. This sensor offers a maximum ranging frequency of \qty{15}{\hertz} at an $8\times8$ pixel resolution, covering a FoV of \qty{45}{\degree}. Moreover, the VL53L5CX provides a pixel validity matrix alongside the 64 depth measurements, automatically identifying and flagging noisy or out-of-range measurements. To facilitate the utilization of multi-zone ranging sensors on the Crazyflie 2.1 platform, a custom deck was employed specifically to accommodate four VL53L5CX ToF sensors, as depicted in Figure~\ref{fig:hardware-b}. The four ToF sensors face the front, back, left, and right directions, enabling obstacle detection within a cumulative FoV of \qty{180}{\degree}. 
Consequently, the final design of the custom deck weighs a mere \qty{2.96}{\gram}. 
The hardware design is released as open source.

\section{Experimental Results} \label{sec:results}

We demonstrate the efficacy of our system by conducting practical field experiments within both controlled and realistic environments, encompassing general indoor areas. Moreover, C-SLAM is compared with SoA recent works in terms of accuracy and scalability, demonstrating how hundreds of robotic agents can collaborate, sharing data and scans over standard wireless protocols. Within the scope of our mapping mission, a variable number of drones, between one and four, explore different environments, employing our exploration and C-SLAM algorithms. Furthermore, a scalability test varying the swarm size between 2 and 200 agents has been conducted in a separate experiment. During flight, drones utilize UWB radio communication to convey their respective positions to one another, as well as the distances between them. Our evaluation is centered on assessing how the coverage time scales with the number of drones, the absolute trajectory error (ATE) of each drone within the swarm, and the mapping accuracy of the collective global map. 

\subsection{Experiments in a controlled environment}

\begin{figure} [t]
\begin{centering}
\includegraphics[width=\linewidth]{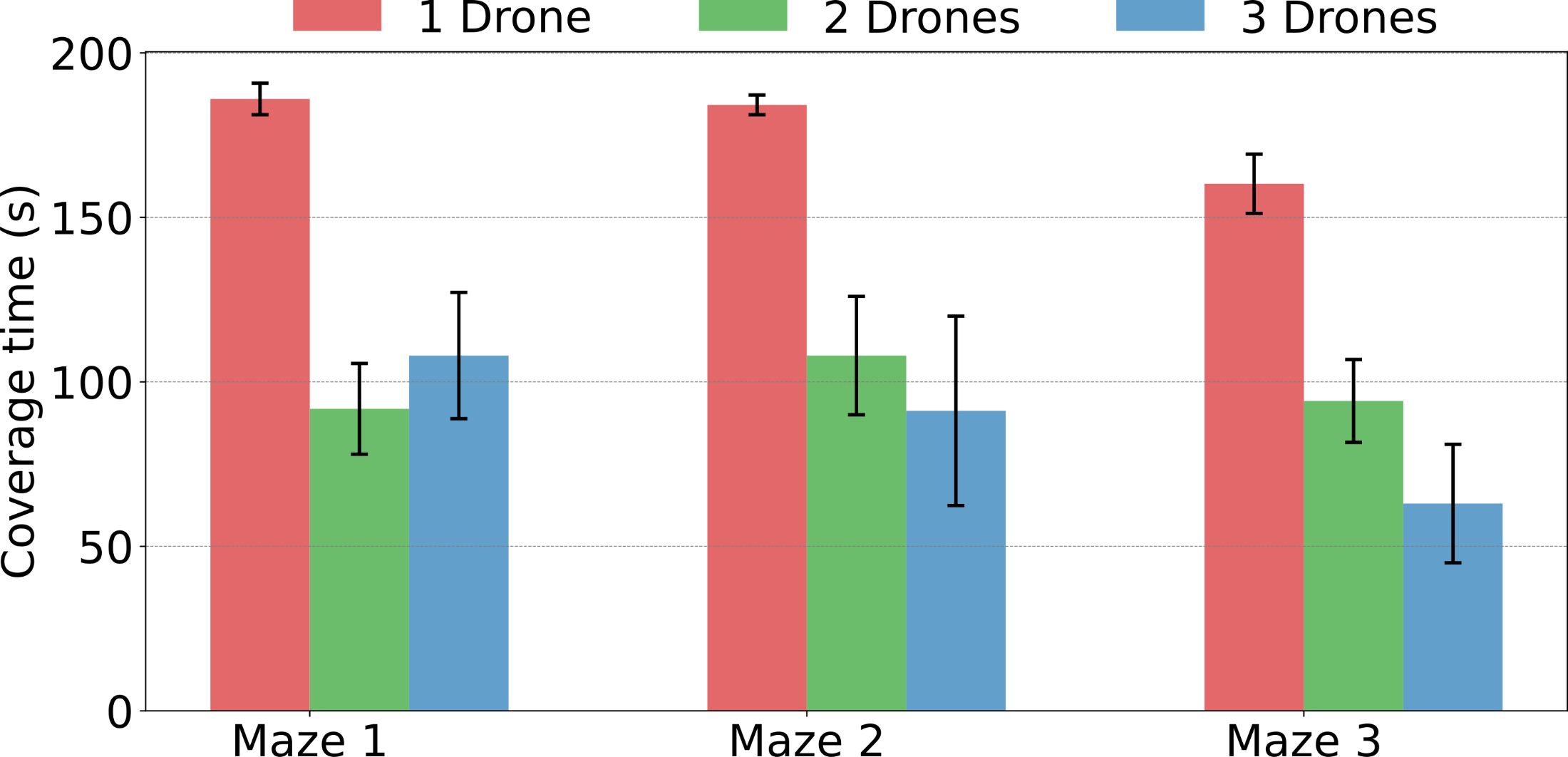}
\par\end{centering}
\centering{}
\caption{The coverage time for one, two, and three drones, illustrating the effectiveness of collaborative exploration.}
\label{fig:coverage}
\end{figure}

\begin{figure*} [t]
\begin{centering}
\includegraphics[width=\linewidth]{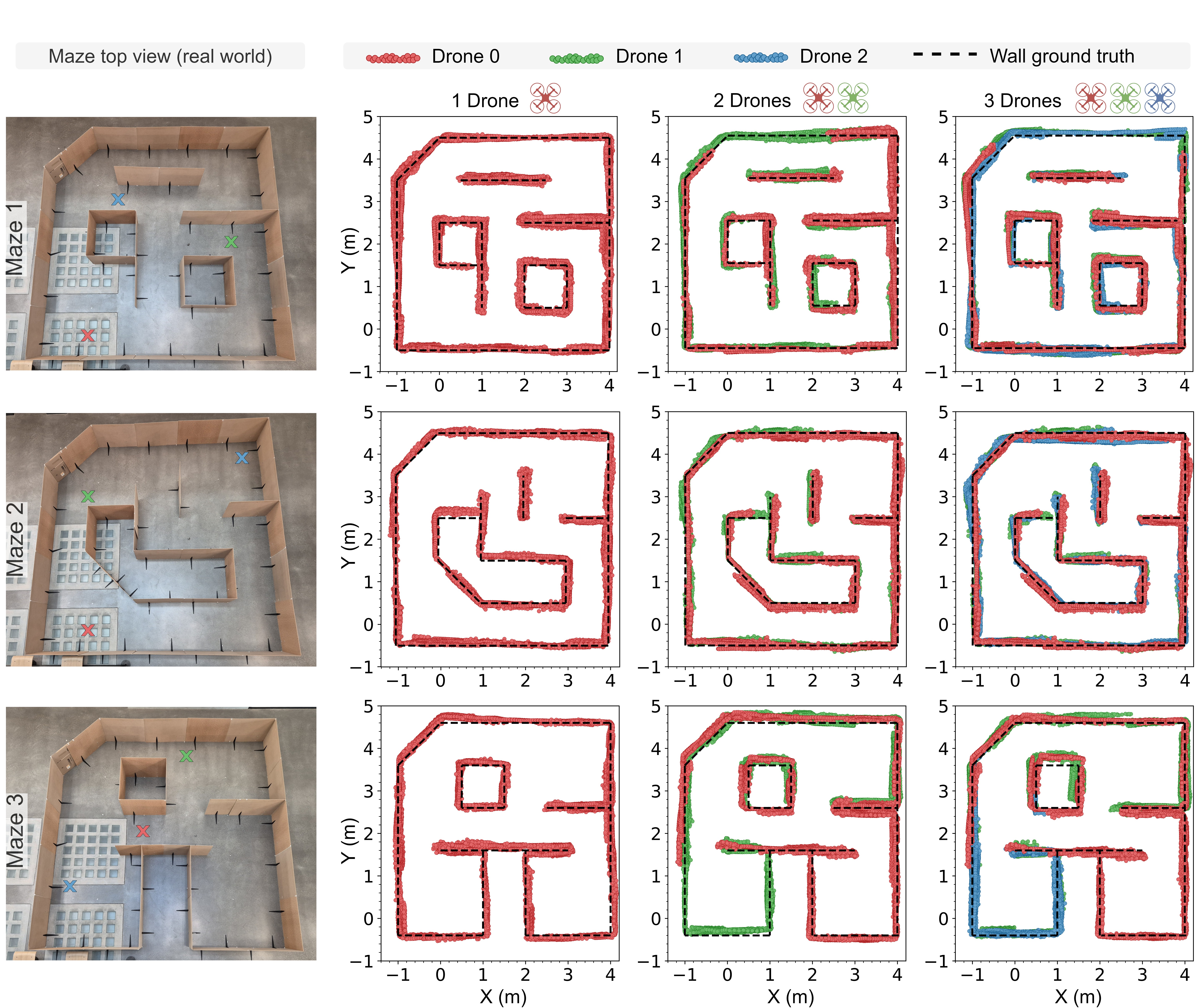}
\par\end{centering}
\centering{}
\caption{Mapping results in a controlled environment. The first column illustrates top view photos of each maze, taken in-field. The other columns show the maps produced at the end of the mission in each maze for one, two, and three drones, illustrating how the individual maps are aligned and merged by our C-SLAM algorithm. }
\label{fig:results}
\end{figure*}

First, we evaluate our system in a controlled environment consisting of mazes built out of chipboard panels of $\qty{1}{\meter} \times \qty{0.8}{\meter}$. The goal is to evaluate the coverage time and mapping accuracy, observing how these metrics are influenced by varying the number of drones. 
To demonstrate the adaptability of our system to different spatial configurations, we evaluated its mapping efficacy across several mazes with distinct geometries. The Vicon Vero 2.2 motion capture system installed in our testing arena provides the ground truth for the evaluation. Figure~\ref{fig:results} illustrates the three distinct mazes subjected to this analysis, with the take-off positions of the drones marked by an “×” of three distinct colors. After take-off, the drones embark on their exploration and mapping tasks, propelled by the underlying exploration algorithm. For each configuration, Figure~\ref{fig:coverage} shows the time the drones require to cover all accessible areas within the maze. As anticipated, the longest duration correlates with the experiments utilizing a single drone. Moreover, introducing a second drone yields a significant reduction in coverage time by 48\%, 43\%, and 41\% for Maze 1, Maze 2, and Maze 3, respectively. 
However, transitioning to a configuration with three drones does not yield a proportional decrease in coverage time; reductions are observed at only 50\% and 60\% for Maze 2 and Maze 3, respectively, when compared to the single-drone setup. This suggests a saturation trend when increasing the number of agents per square meter. For Maze 1, the coverage time increases by 18\% compared to the two-drones experiment. This phenomenon can be attributed to the drones intersecting paths twice in Maze 1, as opposed to once and not at all in Mazes 2 and 3, respectively. Since intersections compel the drones to reverse direction and retrace their previously covered path, they increase the coverage time. This highlights the importance of adjusting the drone count based on the size of the explorable area.

\begin{table} [b]
\centering
\begin{tabular}{cc|cc|ccc} 
\hline\hline
       & 1 Drone                                 & \multicolumn{2}{c|}{2 Drones}                                                   & \multicolumn{3}{c}{3 Drones}                                                                                               \\
       & \textcolor[rgb]{0.886,0.408,0.412}{\#1} & \textcolor[rgb]{0.886,0.408,0.412}{\#1} & \textcolor[rgb]{0.42,0.741,0.42}{\#2} & \textcolor[rgb]{0.886,0.408,0.412}{\#1} & \textcolor[rgb]{0.42,0.741,0.42}{\#2} & \textcolor[rgb]{0.384,0.627,0.796}{\#3}  \\ 
\hline
Maze 1 & 15.7                                    & 15.9                                    & 13.3                                  & 16.3                                    & 24.0                                  & 19.4                                     \\
Maze 2 & 12.8                                    & 12.5                                    & 16.2                                  & 12.7                                    & 13.4                                  & 17.1                                     \\
Maze 3 & 16.5                                    & 15.0                                    & 21.5                                  & 15.6                                    & 22.7                                  & 13.8                                     \\
\hline\hline
\end{tabular}
\caption{The ATE (in \qty{}{\centi\meter}) for each robot in the swarm, highlighting the positioning precision for each configurations. \label{tab:ate}}
\end{table}

Table~\ref{tab:ate} shows the ATE for each configuration, which varies in the range of \qty{12.5}{\centi\meter} -- \qty{24}{\centi\meter}, with the highest values for the configuration of Maze 1 and three drones. Given that rotational movements worsen the ATE, due to yaw angle observability limitations of nano-UAVs operating indoors, and each encounter between drones necessitates a \qty{180}{\degree} spin, it logically follows those configurations characterized by a higher frequency of intersections exhibit increased ATE values. Furthermore, Maze 3 also shows a relatively high ATE, regardless of the number of drones. In this case, it is not the drone intersection that increases the ATE but the geometry of the maze, as the two ''rooms'' at the bottom are only connected through an upper path. The absence of loop closures in the lower section of Maze 3 results in distortion of the trajectory, thereby influencing the ATE. Within the scope of this work, we specifically selected a constrained drone platform featuring limited odometry capabilities that restrict the trajectory estimation accuracy to demonstrate how C-SLAM provides stability and robustness even under poor observability conditions.

Figure~\ref{fig:results} (right side) presents the mapping outcomes for each maze aimed at evaluating the effect of multiple inter-drone loop closures on the map’s integrity. The metric used to assess the overall C-SLAM accuracy is the absolute mapping error. This metric first calculates the projection to the closest straight line (or its extension) for every point in the map. Then, the mapping error is computed as the root-mean-squared-error (RMSE) of all the projections. The mapping RMSE does not only depend on the state estimation error, but it also correlates with the shape of the environment and the measurement noise of the depth measurements. The overall swarm mapping error spans from \qty{6.4}{\centi\meter} to \qty{12}{\centi\meter}, as shown in Table~\ref{tab:mapping}. 
The results are aligned with Table~\ref{tab:ate}, yielding the smallest overall mapping error for Maze 2, and the largest error for Maze 3. This is expected because each drone has its own biases, and anchoring the maps in a finite number of points may not completely align them. 

\begin{table} [b]
\centering
\begin{tabular}{cc|c|c} 
\hline\hline
       & 1 Drone & 2 Drones & 3 Drones  \\ 
\hline
Maze 1 & 6.7     & 9.5      & 10.2      \\
Maze 2 & 6.4     & 6.8      & 7.2       \\
Maze 3 & 7.3     & 11.3     & 12.0      \\
\hline\hline
\end{tabular}
\caption{The mapping error (in \qty{}{\centi\meter}) for each configuration. \label{tab:mapping}}
\end{table}

Table~\ref{tab:lc} shows how many loop closures are performed in each mission. While the numbers are comparable among configurations, we notice a larger amount of intra-drone loop closures when mapping Maze 3 with three drones. This is due to the nature of the exploration policy, as once a drone reaches one of the bottom ''rooms'' of Maze 3, it is likely to perform several loops until it exits the area. This leads to multiple intra-drone loop closures performed in a short time. The maps presented in Figure~\ref{fig:results} underwent a filtering phase, where points with a low density of neighbors were removed, assuming they likely represent outliers corresponding to nearby drones rather than real objects. This approach enhances the map's accuracy by ensuring it reflects only points associated with physical features, based on the premise that genuine features are typically surrounded by a cluster of points. 

\begin{table}
\centering
\begin{tabular}{cc|cc|ccc} 
\hline\hline
       & 1 Drone                                 & \multicolumn{2}{c|}{2 Drones}                                                   & \multicolumn{3}{c}{3 Drones}                                                                                               \\
       & \textcolor[rgb]{0.886,0.408,0.412}{\#1} & \textcolor[rgb]{0.886,0.408,0.412}{\#1} & \textcolor[rgb]{0.42,0.741,0.42}{\#2} & \textcolor[rgb]{0.886,0.408,0.412}{\#1} & \textcolor[rgb]{0.42,0.741,0.42}{\#2} & \textcolor[rgb]{0.384,0.627,0.796}{\#3}  \\ 
\hline
Maze 1 & 6                                       & 8                                       & 4 (\textbf{3})                        & 3                                       & 2 (\textbf{6})                        & 4 (\textbf{6})                           \\
Maze 2 & 7                                       & 7                                       & 8 (\textbf{5})                        & 7                                       & 8 (\textbf{5})                        & 7 (\textbf{5})                           \\
Maze 3 & 4                                       & 9                                       & 4 (\textbf{4})                        & 9                                       & 7 (\textbf{3})                        & 15 (\textbf{3})                          \\
\hline\hline
\end{tabular}
\caption{Number of intra-drone loop closures per experiment. Inter-drone loop closures are shown in bold. \label{tab:lc}}
\end{table}

\subsection{Experiments in a Real-World Indoor Environment} \label{sec:results_real_env}

\begin{figure} [t]
\begin{centering}
\includegraphics[width=\linewidth]{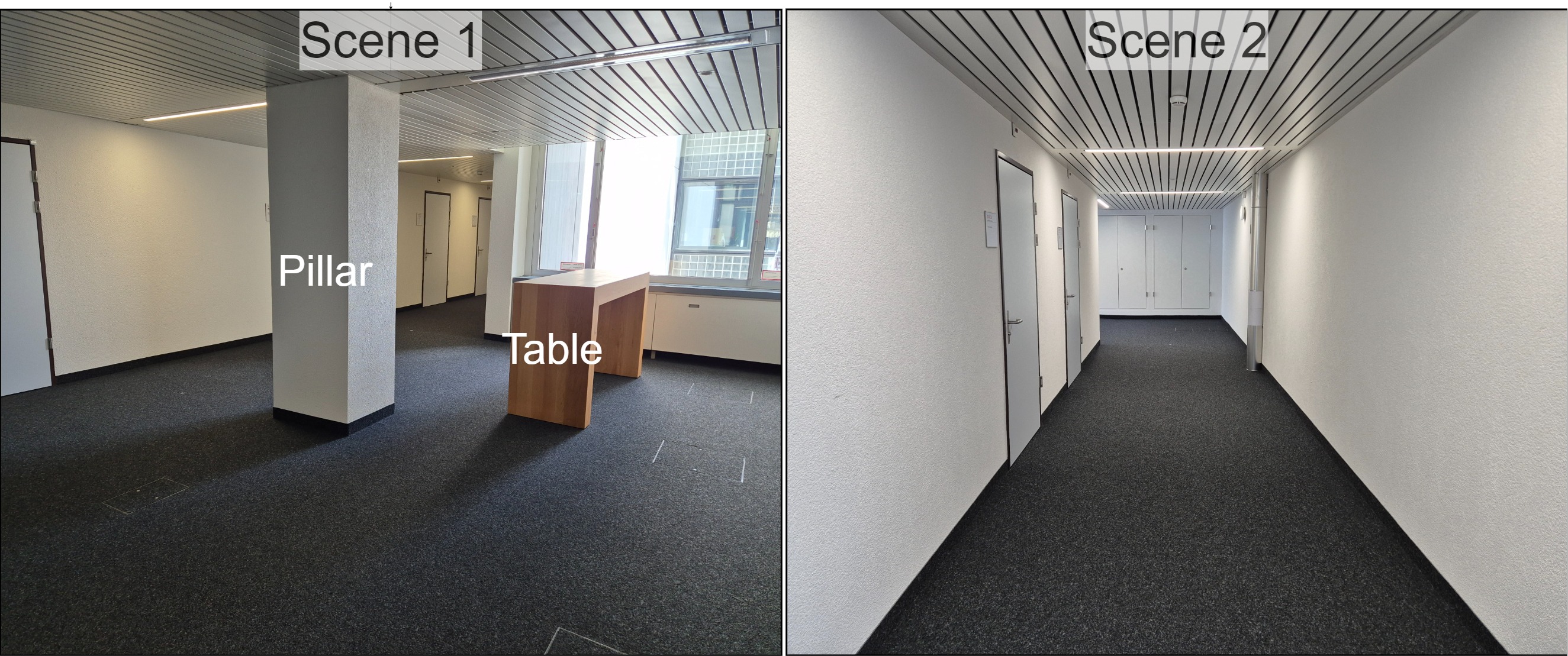}
\par\end{centering}
\centering{}
\caption{Samples from the real environment where the mapping mission is conducted, highlighting the encountered obstacles.}
\label{fig:real_footage}
\end{figure}

\begin{figure} [b!]
\begin{centering}
\includegraphics[width=\linewidth]{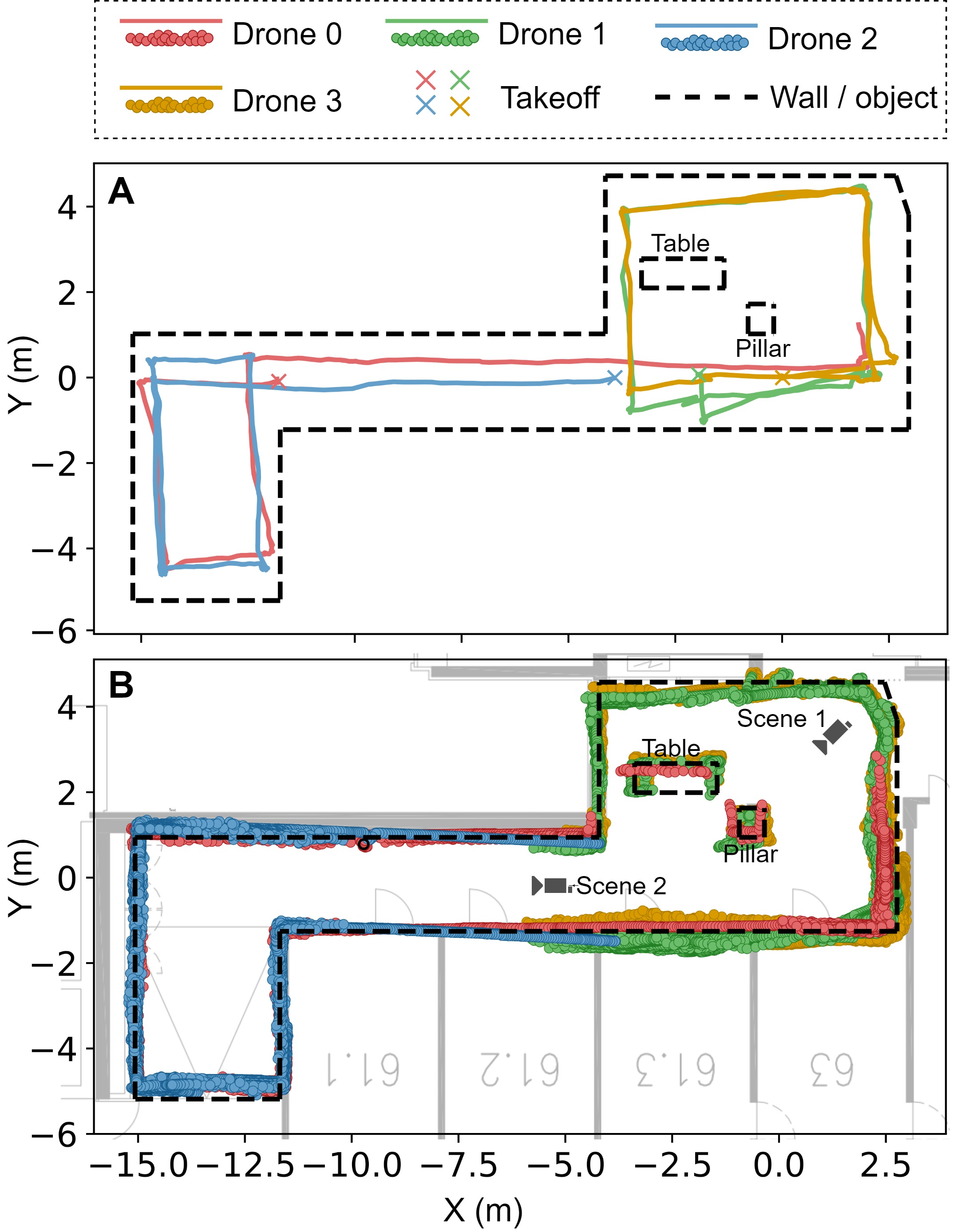}
\par\end{centering}
\centering{}
\caption{Mapping results in a real world environment. (A) The trajectories of each drone, color-coded according to the legend for clear differentiation (B) The collaborative map created by the drone swarm}
\label{fig:results_real}
\end{figure}

We also prove the effectiveness of our C-SLAM in a realistic environment. For this purpose, we map an \qty{18}{\meter} $\times$ \qty{10}{\meter} area in an office environment employing a swarm of four nano-UAVs, depicted in Figure~\ref{fig:hardware}-(a). The number of drones was estimated from the results in Figure~\ref{fig:coverage}, where coverage time saturation implies an optimal number of agents per square meter. Moreover, the experiment setup is comparable with recent collaborative mapping papers~\cite{tian2022kimera}. Two scenes of the environment are illustrated in Figure~\ref{fig:real_footage}, while the obstacles are labeled in Scene 1. This environment is selected as the worst-case scenario, with few favorable areas for loop closure and a thick concrete wall affecting wireless communication. The proposed experimental setup pushes the limits of C-SLAM, enabling distributed mapping on ultra-constrained nano-robotic platforms. Since there is no trajectory ground truth in this experiment, we represent in Figure~\ref{fig:results_real}-A the onboard estimated trajectories that were corrected by the C-SLAM algorithm to show the effectiveness of our exploration algorithm in distributing the drones through the whole environment. They initiate their flight from positions denoted by '×', achieving complete area coverage in approximately \qty{60}{\second}. 

Figure~\ref{fig:results_real}-B shows how the map looks at the end of the mission, where the data points produced by each drone are represented with a distinct color. Additionally, the locations where the two environmental snapshots were captured are indicated. Notably, the experiment exhibits accurate corner alignment attributed to inter-drone loop closures. The most significant discrepancy is noted adjacent to the longest wall at the bottom of the area, approximately \qty{15}{\meter}, where the absence of loop closures over extended distances results in substantial accumulated trajectory drift, thereby inducing ATE errors that are not fully rectifiable. Conclusively, the experiment leads to a mapping error of \qty{29.7}{\centi\meter}. It is important to note that, in contrast to the maze-based experiments characterized by a more confined area, in this experiment, the drones frequently encountered UWB radio out-of-range situations limiting the relative intra-swarm localization, predominantly due to the obstructions of concrete walls found in the line of sight. 
Despite these challenges, our system demonstrated robustness against connection-drop conditions, and it consistently demonstrated the ability to recover from such states when the drones returned within the communication range.

\subsection{Scalability Study}

\begin{figure}[t]
    \centering
    \begin{subfigure}{0.875\columnwidth}
        \centering
        \hspace{-0.5cm}
        \includegraphics[width=\linewidth]{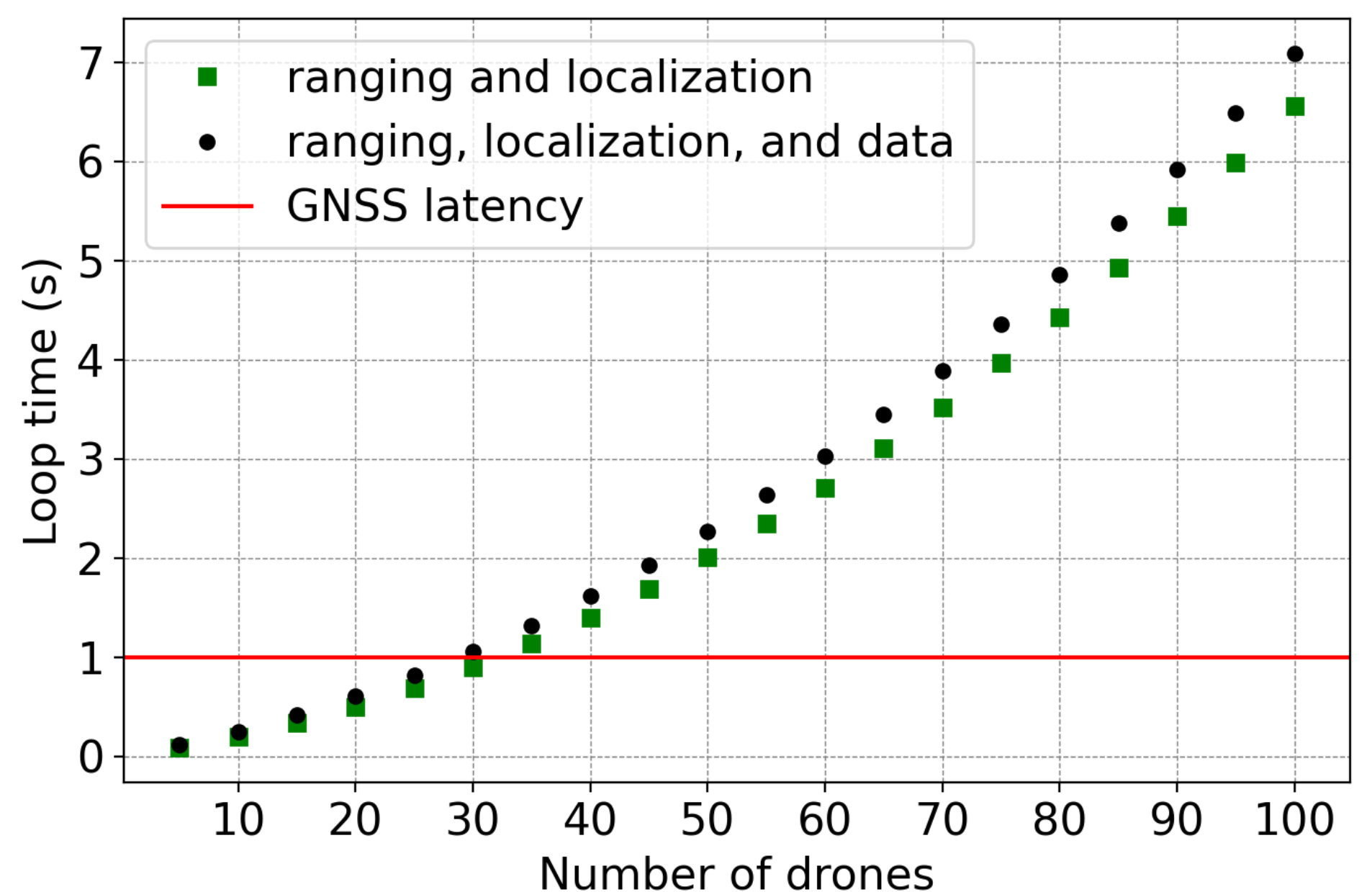}
        \caption{Loop time.
        \label{fig:scalability0}}
    \end{subfigure}
    
    \begin{subfigure}{\columnwidth}
        \centering
        \includegraphics[width=\linewidth]{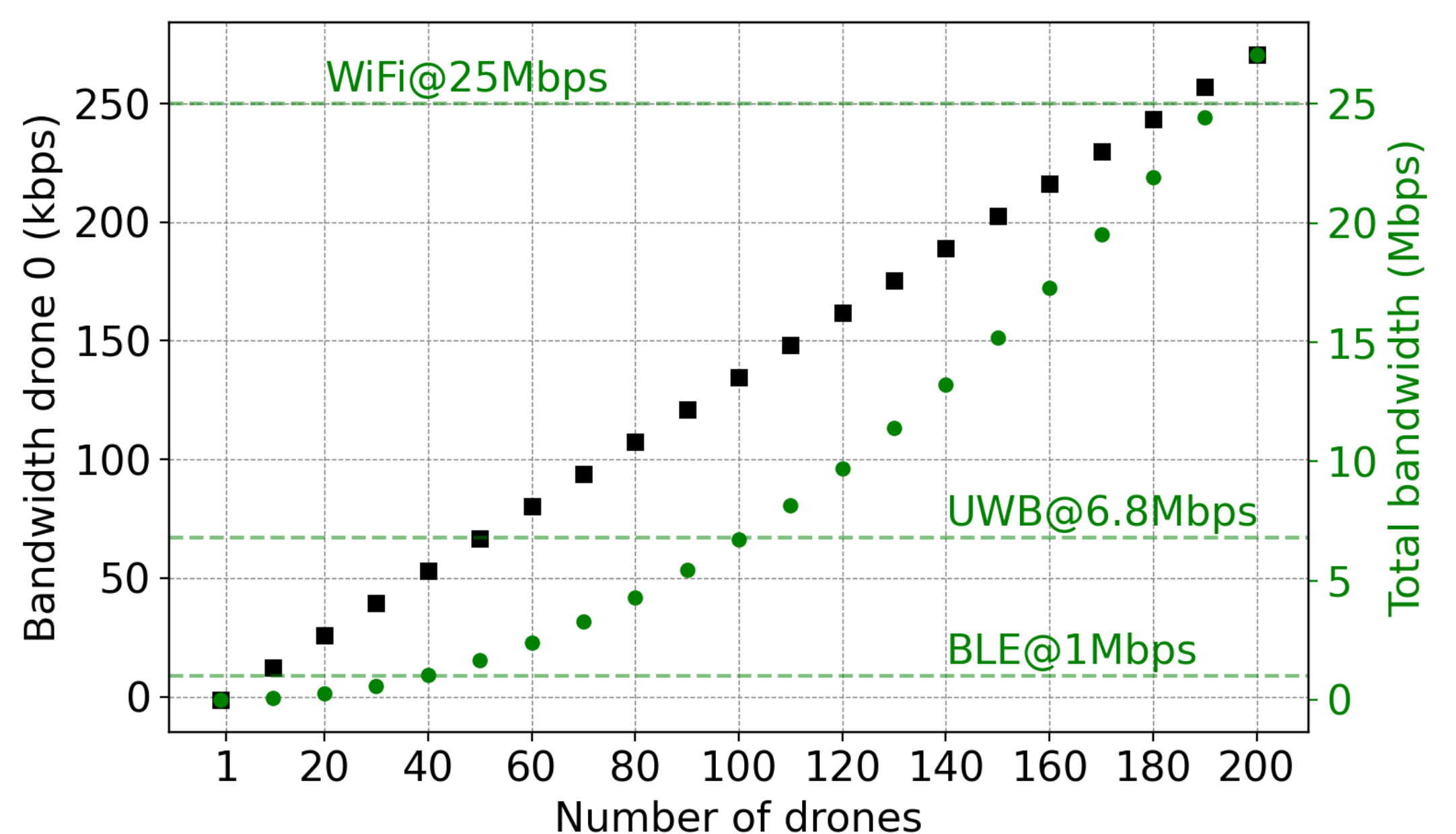}
        \caption{Necessary radio bandwidth.
        \label{fig:scalability1}}
    \end{subfigure}
    \caption{Scalability study. (a) The loop time, including UWB ranging and localization alone and with the effect of transferring scans. (b) The radio bandwidth per drone (drone 0 has the highest bandwidth) and for the whole swarm, highlighting the nominal data rate of common wireless technologies}
    \label{fig:edo-err}
\end{figure}

Our system prioritizes scalability, hardware cost, and computational load. A scalability study, based on the intra-swarm data communication and localization, is presented in Figure~\ref{fig:scalability0} and Figure~\ref{fig:scalability1}, demonstrating the possibility of supporting up to 200 robots. 
We define loop time as the total time necessary for the drones in the swarm to communicate their positions to each other, the distances between them, and the scans. In the communication scheme we propose, only one drone transmits at a time, and the transmission scheduling for each drone occurs in increasing order based on its ID. Therefore, a large loop time would impact the update rate of the positions, decreasing the robustness of collision avoidance. Figure~\ref{fig:scalability0} shows how the loop time depends on the number of drones. The position and distance updates are relevant for inter-drone collision avoidance, while the scan transfer is associated with mapping and loop closure. 
Consequently, Figure~\ref{fig:scalability0} illustrates the loop time with and without considering the influence of the scans. Given that the scan acquisition frequency is usually considerably lower than the loop frequency, this study presumes that each drone sends at most one scan per loop and no more than 20\% of the drones in the swarm send a scan within the same loop. We highlight that a latency comparable to GNSS-based localization is obtained for a swarm size of 30 drones. Moreover, while previous works~\cite{huang2022edge} achieve a \qty{2.7}{\second} latency with a swarm of 10 drones (data transmission only), our system allows a swarm of 55 robots with the same latency (data transmission and localization).

We highlight that the loop time in Figure~\ref{fig:scalability0} mainly depends on transmitting the drones’ positions and distances. However, combining an absolute positioning system with our C-SLAM pipeline would alleviate the need to wirelessly measure the relative drone positions. In such a scenario, the only scalability limitation would lie in the necessary bandwidth for transferring the scans. Figure~\ref{fig:scalability1} shows the bandwidth requirement as a function of the number of drones. Furthermore, we represent both the total swarm bandwidth and the bandwidth required by drone 0 (i.e., the worst case). In addition, the figure shows that conventional communication technologies~\cite{jamshed2022challenges} such as WiFi, UWB, and BLE can accommodate about 190, 100, and 40 drones, respectively. Although the WiFi protocol (Wi‑Fi 6 - IEEE 802.11ax) can reach up to \qty{600}{Mbps}, in line with the scope of this work, we selected a cheap and miniaturized WiFi module (NINA-W10 from u-blox) supporting up to \qty{25}{Mbps}. Within this investigation, we consider an average number of five scans per minute and a scan size of \qty{2}{\kilo\byte}. The possibility of pairing our setup with BLE enables even teams of insect-size robots to collaboratively map a previously unknown environment~\cite{iyer2020wireless}.

\subsection{Onboard Execution Time and Energy Analysis}

We provide an analysis of the execution time of the adopted algorithms. Our focus lies on evaluating ICP and PGO onboard the robotic platform and only relying on its computing capabilities. Table~\ref{tab:analysis} (top) presents the execution time of ICP as a function of the scan size. We sweep the scan size in the range of 128 – 1024 points and obtain an execution time in the range of \qty{3}{\milli\second} -- \qty{138}{\milli\second}. In this work, we select a scan size of 640, as the best trade-off between scan-matching accuracy and execution time. Note that the average power consumption (in the range of \qty{122}{\milli\watt} -- \qty{178}{\milli\watt}) increases with the scan size, because a larger scan results in a longer activity of the GAP9 CL. Our C-SLAM algorithm optimizes the trajectory of each individual robot relying on the graph-based SLAM proposed in~\cite{niculescu2023nanoslam}. This algorithm uses a hierarchical approach, which firstly splits the graph into multiple subgraphs, and then optimizes each subgraph individually. Since the memory constraints limit the maximum size of the graph to about 440 poses when performing graph optimization, the hierarchical method allows to optimize much larger graphs (up to about 3000 poses). The subgraph split is typically performed using a distance threshold, and each subgraph corresponds to a travelled distance of about \qty{0.8}{\meter}. The optimization itself has a quadratic complexity, but since the size of the subgraphs is approximately constant, the execution time of the whole hierarchical optimization is linear in the number of subgraphs, and therefore in the total number of poses (assuming an equal number of poses per subgraph). This is proven by the execution time measurements shown in Table~\ref{tab:analysis} (bottom). Furthermore, the average power consumption is approximately constant in all configurations – about \qty{70}{\milli\watt}.

\begin{table} [t]
\centering
\resizebox{\columnwidth}{!}{
\begin{tabular}{ccccccccc} 
\hline\hline
\multicolumn{9}{c}{ICP analysis}                               \\
\hline
Scan size   & 128  & 256  & 384  & 512  & 640  & 768  & 896  & 1024  \\ 
\hline
Time (ms)   & 3    & 10   & 21   & 36   & 55   & 79   & 107  & 138   \\
Power (mW)  & 122  & 150  & 165  & 170  & 172  & 175  & 176  & 178   \\
Energy (mJ) & 0.54 & 1.8  & 3.8  & 6.5  & 10   & 14   & 19   & 25    \\ 
\hline
\multicolumn{9}{c}{PGO analysis}                               \\
\hline
Poses       & 800  & 1000 & 1200 & 1400 & 1600 & 1800 & 2000 & 2200  \\ 
\hline
Time (ms)   & 206  & 259  & 320  & 374  & 431  & 485  & 534  & 593   \\
Energy (mJ) & 14.4 & 18.1 & 22.4 & 26   & 30.2 & 34   & 38   & 41.5  \\
\hline\hline
\end{tabular}
}
\caption{Execution analysis of execution time, average power and energy of the ICP and graph-based SLAM algorithms onboard the GAP9 SoC. \label{tab:analysis}}
\end{table}

\section{Discussion} \label{sec:discussion}

So far, accurate collaborative SLAM has been a prerogative of powerful and expensive robotic platforms such as GPUs-based ones, limiting swarm formations to few units due to the high volume of exchanged data necessary for loop closure and map alignment~\cite{zhou2022swarm}. Moreover, the majority of existing SLAM systems rely on external infrastructure support for localization and computation~\cite{amala2023drone, rosinol2020kimera}, such as GNSS or motion capture systems. 

Recent SoA works on onboard and decentralized solutions mainly focus on maximizing the mapping accuracy but do not address latency and scalability~\cite{placed2023survey, zhong2023dcl}. Despite relying on high-end platforms, recent works reach an average global map update in the range between \qty{0.1}{\second} to \qty{7}{\second}~\cite{rosinol2020kimera, niculescu2023nanoslam}. Moreover, such a large data volume bounds the minimum latency, which in some cases reaches up to \qty{5}{\second} due to computation and data transmission overhead~\cite{huang2022edge}. Despite these powerful computational platforms, more similar to a server than an embedded system, the CPU load level reaches a mean of 93\% \cite{huang2022edge} or half of the total load for autonomous car competitions \cite{baumann2024forzaeth} only to perform 2D mapping. Under these circumstances, SoA works on distributed and collaborative SLAM~\cite{tian2022kimera} feature a mapping error in the range of \qty{20}{\centi\meter} to \qty{25}{\centi\meter} for indoor operation. We achieved a comparable error of about \qty{30}{\centi\meter}, but our system requires less than \qty{1.5}{\mega\byte} of RAM and a sensor setup that costs about \$20. 

Within our experiments, the drones are aware of their initial take-off locations and continue keeping track of them during the mapping mission. While the initial robot locations might not be known in all real-world applications (e.g., random deployment), our work focuses on the collaboration between drones and maintaining the consistency of the global map over time. To address the initial localization problem, our system can be paired with one of the previous works addressing this issue. For example, the authors in~\cite{schindler2023relative} propose a relative localization algorithm based on pair-wise UWB range measurements.

\subsection{Maximum Mappable Area}
The size of the pose graph each drone stores and optimizes depends on the distance it traveled (i.e., internal poses). However, the memory footprint PGO requires depends on both nodes and edges in the graph, the latter being impacted by the number of loop closures. Thus, the maximum graph size that can be optimized depends on the size of the environment and its geometry - paired with how many loop closures arise. For example, an environment where a robot acquires 1400 poses and performs five loop closures results in $\sim$\qty{250}{\milli\second} execution time onboard a GAP9 SoC. The largest graph an individual drone can optimize in similar conditions has about 3000 poses, within the \qty{1.5}{MB} constraint, corresponding to an environment of about \qty{400}{\meter\squared}. Due to the distributed nature of our C-SLAM solution, this area scales about linearly with the number of robots.

\subsection{Comparison with LiDAR-Based Mapping}

\begin{figure} [t]
\begin{centering}
\includegraphics[width=\linewidth]{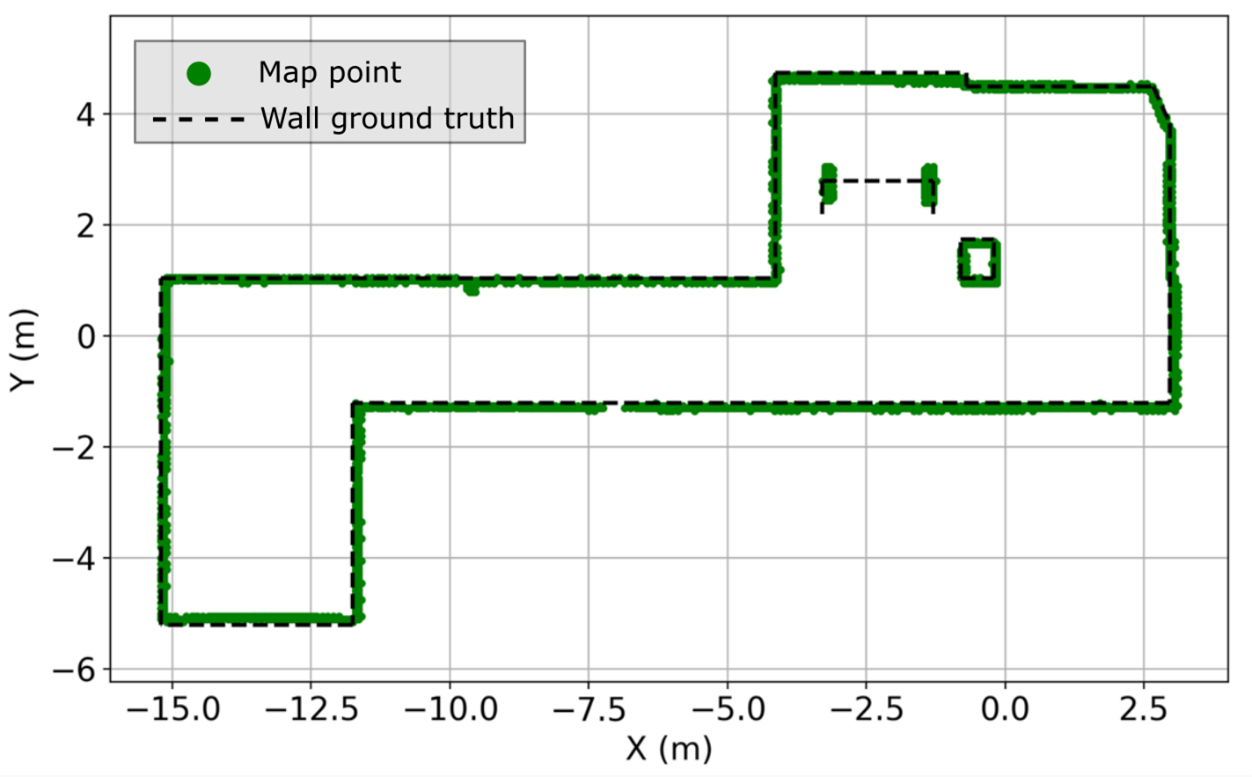}
\par\end{centering}
\centering{}
\caption{The map is obtained by mapping the same environment from Section~\ref{sec:results_real_env} with a Unitree A1 Robot Dog equipped with a LiDAR and using Cartographer as SLAM engine.}
\label{fig:robodog}
\end{figure}

Although the ToF sensors we employ in this work are lightweight, energy-efficient, and cost-effective, conventional LiDARs maintain superiority in terms of accuracy. Primarily, LiDARs commonly feature an extended operational range, facilitating the mapping of distant environmental regions. Conversely, our ToF depth sensor is constrained by a maximum range of \qty{4}{\meter}. Moreover, LiDARs frequently boast superior angular resolution, leading to measurement precision that is less dependent on distance magnitude, allowing for higher accuracy scans and scan-matching. However, to precisely quantify the differences in mapping accuracy, we map again the environment from Figure~\ref{fig:real_footage} using a ground legged robot (i.e., Unitree A1) equipped with Hokuyo UTM-30LX-EW 2D LiDAR configured at \qty{20}{\hertz}. The processing is conducted on an Intel NUC10i7FNKN with $10^{th}$ gen 6-core i7 CPU, alongside a NVIDIA Jetson Xavier NX. In this configuration, the computing power is \qty{45}{\watt}, total RAM memory \qty{32}{\giga\byte}, and the computing cost in the range of \$1500 USD. In contrast, our collaborative SLAM requires orders of magnitude lower hardware specifications, ~\qty{100}{\milli\watt}, \qty{1.5}{\mega\byte}, ~\$10 USD, respectively. The same applies to the sensor platform. 

Figure~\ref{fig:robodog} shows the map obtained with the Robot Dog-based system and Cartographer~\footnote{\url{github.com/cartographer-project/cartographer}}, a SoA solution for real-time SLAM. Note that the dashed ground truth line, which is not mapped, corresponds to the table plate located at a height of \qty{1}{\meter}, outside the LiDAR's field of view. Despite the strong difference in computational and sensing power, the resulted map has an accuracy of \qty{10.05}{\centi\meter}, which is in the same order of magnitude as the result achieved by our mapping system (i.e., \qty{29.7}{\centi\meter}). 
However, the accuracy difference is mainly impacted by the lower precision of the optical flow-based odometry used by the nano-UAVs, which is inferior to the odometry of the ground robots. More in detail, the legged robot’s odometry accuracy is mainly impacted by encoders, while for the optical-flow-based odometry used by our nano-UAVs, the errors depend on multiple factors such as illumination, reflections, texture, which vary within the testing area. These errors are therefore not deterministic and more difficult to correct or model.

\subsection{Enabling mapping in 3D}

\begin{figure} [t]
\begin{centering}
\includegraphics[width=\linewidth]{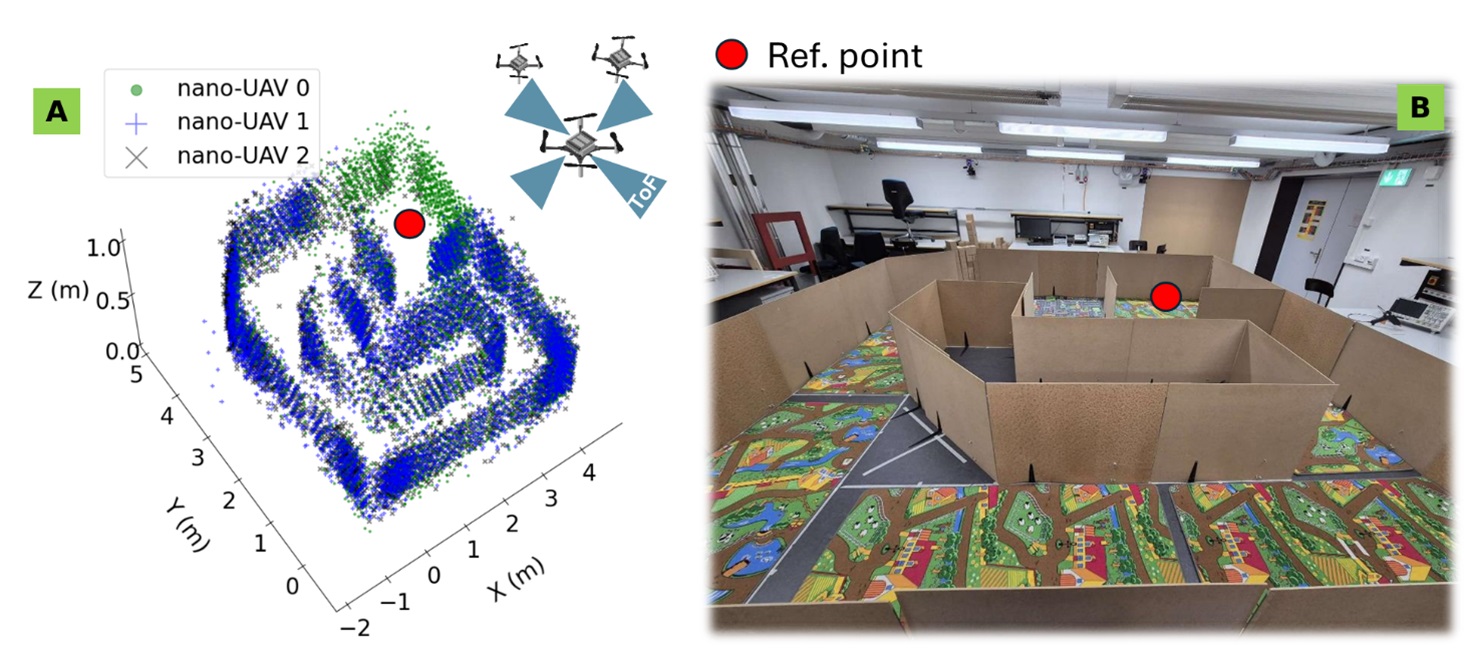}
\par\end{centering}
\centering{}
\caption{(A) The generated 3D map acquired from three nano-UAVs. (B) The environment mapped in (A).}
\label{fig:3dmap}
\end{figure}

The C-SLAM pipeline we propose supports only 2D mapping and 2D loop closure, an optimal tradeoff between mapping accuracy and memory footprint. However, the raw depth measurements from the depth sensors are acquired in matrix form, with a total of 64 pixels distributed in a resolution of $8\times8$. While our 2D approach reduces the measurements from matrix to row format, a 3D map can be obtained by projecting all matrix measurements in the world frame. In this setup, the loop closure and ICP are still performed in 2D, while the generated graph map will result in a 3D perspective. An example is provided in Figure~\ref{fig:3dmap}, where a maze environment mapped concurrently with three nano-UAVs and its respective 3D map are depicted. In this configuration, the scan size becomes $6\times$ larger, due to the necessity of storing and transmitting 64 pixels for each sensor.
Table~\ref{tab:frame_structure} shows the structure of a depth frame for both scenarios and how the dimension changes when performing 3D mapping compared to the 2D scenario.

\begin{table}
\centering
\begin{tabular}{lcc|cc} 
\hline\hline
               & \multicolumn{2}{c}{2D Mapping} & \multicolumn{2}{c}{3D Mapping}   \\
          & Format                 & Size  & Format                  & Size   \\ 
\hline
Pose ID        & \textit{int32}         & \qty{4}{\byte}   & \textit{int32}          & \qty{4}{\byte}    \\
Timestamp      & \textit{int32}         & \qty{4}{\byte}   & \textit{int32}          & \qty{4}{\byte}    \\
Pose           & 3 $\times$ \textit{float32}   & \qty{12}{\byte}  & 3 $\times$ \textit{float32}    & \qty{12}{\byte}   \\
Depth          & 4 $\times$ 8 $\times$ \textit{int16} & \qty{64}{\byte}  & 4 $\times$ 64 $\times$ \textit{int16} & \qty{512}{\byte}  \\ 
\hline
\textbf{Total} &                        & \textbf{\qty{84}{\byte}}  &                         & \textbf{\qty{532}{\byte}}  \\
\hline\hline
\end{tabular}
\caption{Memory structure of a depth frame. \label{tab:frame_structure}}
\end{table}
\section{Conclusions} \label{sec:conclusions}

This paper introduced a decentralized and lightweight collaborative SLAM approach, tailored for low-cost robotic platforms, including those with extremely limited hardware resources. Our solution supports the coordination of large robot swarms, effectively managing the complexities associated with scalability by optimizing the data traffic and the exploration strategy.
Moreover, by reducing the sensing and computation costs to approximately \$20 per unit, our solution offers a cost-effective alternative to traditional high-end SLAM systems, opening up new possibilities for deploying SLAM in budget-constrained applications. To efficiently manage the computation required for SLAM, our system distributes the intensive tasks across RISC-V parallel low-power platforms, operating onboard each individual robot in real-time.
Furthermore, the collaborative exploration strategy that we propose decreases the coverage time while ensuring accurate mapping.

We evaluated our solution in various environments using a swarm of centimeter-size UAVs weighing only \qty{46}{\gram}, achieving a mapping accuracy below \qty{30}{\centi\meter}. This level of precision is comparable to high-end SoA solutions, yet it comes with drastically reduced cost, memory, and computational requirements.

\section*{Acknowledgments}
Authors thank Davide Plozza and Cristian Cioflan for their support during field tests. The GAP9 Deck V2 is designed by Victor Javier Kartsch Morinigo and Hanna Müller with the support of Greenwaves Technologies.

\bibliographystyle{IEEEtran}
% \bibliography{IEEEabrv,bibliography}


\end{document}